\documentclass[10pt, a4paper]{article}         
\usepackage[top=3.6cm, bottom=3.2cm, left=2.5cm, right=2.5cm]{geometry}

\usepackage{color}
\usepackage{graphicx}
\usepackage{cite}
\usepackage[cmex10]{amsmath}
\usepackage{amssymb}
\usepackage{rotating}
\usepackage{multirow}
\usepackage{makecell}

\usepackage{dsfont}
\usepackage{amsthm}
\usepackage{mathrsfs}
\usepackage{algorithm}
\usepackage{algpseudocode}
\usepackage{subcaption}
\newtheorem{proposition}{Proposition}[section]

\newcommand{\TabHeight}[1]{
	\begin{tabular}{@{}c@{\hspace{1mm}}c@{\hspace{1mm}}c@{\hspace{1mm}}c@{\hspace{1mm}}c@{\hspace{1mm}}c@{\hspace{1mm}}@{}}
		#1
	\end{tabular}
}

\usepackage{xspace}

\makeatletter

\DeclareRobustCommand\onedot{\futurelet\@let@token\@onedot}
\def\@onedot{\ifx\@let@token.\else.\null\fi\xspace}

\def\ie{\emph{i.e}\onedot}

\def\etal{\emph{et al}\onedot}
\makeatother

\begin{document}

\title{Texture Mixing by Interpolating \\Deep Statistics via Gaussian Models}

\author{Zi-Ming~Wang$^1$, Gui-Song~Xia$^{1}$, Yi-Peng~Zhang$^{2}$\\
\\
$^1${\em State Key Lab. LIESMARS, Wuhan University, Wuhan, China.}\\
$^2${\em EECS Depart., Syracuse University, Syracuse, USA}\\
}

\maketitle

\begin{abstract}
Recently, enthusiastic studies have devoted to texture synthesis using deep neural networks, because these networks excel at handling complex patterns in images. In these models, second-order statistics, such as Gram matrix, are used to describe textures. Despite the fact that these model have achieved promising results, the structure of their parametric space is still unclear, consequently, it is difficult to use them to mix textures. This paper addresses the texture mixing problem by using a Gaussian scheme to interpolate deep statistics computed from deep neural networks. More precisely, we first reveal that the statistics used in existing deep models can be unified using a stationary Gaussian scheme.
We then present a novel algorithm to mix these statistics by interpolating between Gaussian models using optimal transport.
We further apply our scheme to Neural Style Transfer, where we can create mixed styles. The experiments demonstrate that our method can achieve state-of-the-art results\footnote{Experiments are available at \url{http://captain.whu.edu.cn/TexMixDeepG}.}.
Because all the computations are implemented in closed forms, our mixing algorithm adds only negligible time to the original texture synthesis procedure.
\end{abstract}
%
%
%

\section{Introduction}
Texture mixing is the process of generating new texture images that possess \emph{averaged} visual characteristics of a given set of exemplars~\cite{peyre2010texture,rabin2011wasserstein,FerradansXPA13,darabi2012image,xia2014synthesizing}. It can provide visually pleasing interpolations of difference textures, therefore, has numerous applications in computer vision and graphics~\cite{risser2010synthesizing,darabi2012image}. Besides, the ability to create smooth morphing textures between textures is regard as an criteria for ``good'' texture synthesis algorithms~\cite{jetchev2016texture}~\cite{wei2009state}.

In the sense that a texture can be modeled by a set of statistics depicting the visual properties of its samples~\cite{julesz1981textons,zhu1998filters,portilla2000parametric}, texture mixing involves ``averaging'' the corresponding set of statistical measures. For copy-based texture synthesis methods, e.g.~\cite{efros1999texture, wei2000fast}, textures can be mixed by combining pixels from multiple inputs by using well-designed procedures such as in~\cite{darabi2012image} or the {\em patch match} scheme~\cite{ruiters2010patch}. These methods handle complex and geometric textures satisfactorily, but they tend to produce verbatim patterns and it is not easy to understand the mixing process. In contrast, statistical parametric texture methods, e.g.~\cite{portilla2000parametric, heeger1995pyramid, galerne2011random} are more principled, and their parameters are better understood, although they are often not as good at handling structured textures. Moreover, with parametric texture models, the mixing of textures can be computed feasibly and more easily by ``averaging'' the corresponding set of parameters, see e.g.~ \cite{xia2014synthesizing, peyre2010texture, rabin2011wasserstein, bar2001texture}.

\begin{figure}[htb!]
\centering
		\includegraphics[width=0.5\linewidth]{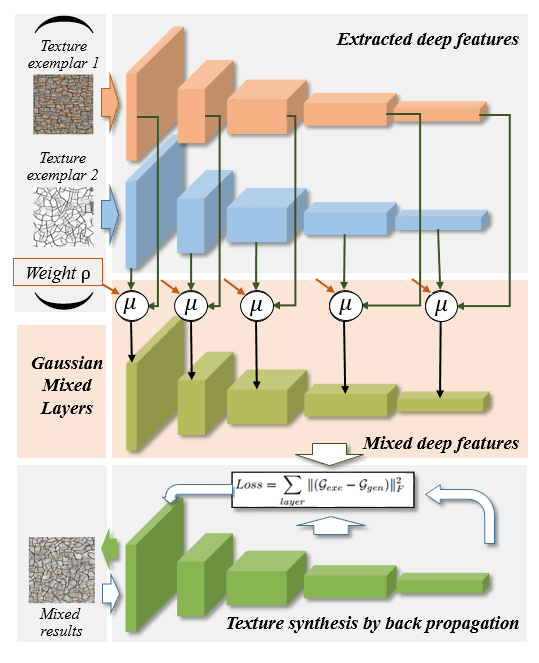}
\caption{The proposed texture mixing scheme. (Top) Two texture exemplars are passed through the CNN. (Middle) The outputs of selected layers are mixed using a Gaussian model. (Bottom) The Gram matrices of the mixed outputs are used as constrains to generate mixed textures.}
\label{Flow}
\end{figure}

A recent breakthrough in texture modelling involves the use of deep convolutional neural networks (CNNs)~\cite{gatys2015texture,liu2016texture,ulyanov2017improved,sendik2017deep,li2017diversified} for texture representation and extends the parametric framework of Portilla and Simoncelli~\cite{portilla2000parametric}. This approach enables us, using parametric models, to synthesize comparable or better textures containing complex patterns than copy-based methods. Under this deep-CNNs framework, researchers also cast the the problem of style transfer into texture transfer~\cite{gatys2016image,ulyanov2017improved}. However, due to the complex structure of the parametric space of deep CNNs~\cite{gatys2015texture,gatys2016image,sendik2017deep}, how to mix textures or styles with these models is still unclear.

In this paper, we address the problem of mixing textures using deep CNNs.
More precisely, after studying existing texture synthesis methods with deep CNNs~\cite{gatys2015texture,liu2016texture,ulyanov2017improved,sendik2017deep,li2017diversified}, we discover that the second-order statistics (e.g.\ Gram matrix and correlation matrix and their variants) used in these methods can be represented as continuous functions of a stationary Gaussian model, so the mixing of these statistics is reduced to the interpolation of Gaussian models.
Therefore, we present a scheme illustrated in Fig~\ref{Flow}, for mixing the statistics of deep CNNs using an optimal transport interpolation of Gaussian models, which enables us to mix textures simply and efficiently. We further apply our scheme to neural-style morphing, where we can interpolate between different styles. We also demonstrate that our mixing algorithm is fully compatible with feed-forward CNNs~\cite{johnson2016perceptual} and instance normalization~\cite{ulyanov2017improved}:  generating mixed textures or stylish photos with mixed styles in a fast forward pass.
Experiments demonstrate that our method achieve state-of-the-art results.
It is also worth noticing that our mixing algorithm adds only negligible time to the original texture synthesis procedure, thanks to the fact that all the mixing computations are in closed forms in Fourier domain.

The rest of this paper is organized as follows: Section~\ref{relatedwork} briefly reviews the related work, Section~\ref{problem} formulates the texture mixing problem, Section~\ref{proposed_method} presents the proposed scheme for mixing deep statistics with Gaussian models, and Section~\ref{implementations} provides all the implementation details. Section~\ref{experiments} compares the proposed methods with state-of-the-arts and analyzes the experimental results. Section~\ref{conclusion} finally draws some conclusion remarks.

\section{Related Work}
\label{relatedwork}
Exemplar-based texture synthesis is the basis of our work, the goal of which is to generate new texture samples from a given texture exemplar~\cite{RaadDDM17}. Work on texture synthesis can be roughly categorized into non-parametric models such as copy-based (also known as {\em patch-based}) methods, see e.g.~\cite{efros1999texture} and statistic parametric models, see e.g.~\cite{portilla2000parametric}. Patch-based models copy pixels or patches directly from texture exemplars to synthesize new texture samples~\cite{efros1999texture,wei2000fast}. This approach is efficient but sometimes generates verbatim patterns, {\em i.e.} precisely the same parts of the exemplar are highly reused in the results. In contrast, statistical parametric methods aim to find a parametric representation of textures, an approach that often allows more control over the synthesis process. Portilla and Simoncelli~\cite{portilla2000parametric} used wavelet and pyramid decomposition to build a parametric texture model, which can synthesize many nature textures, even those containing geometric patterns.
Stationary Gaussian model~\cite{galerne2011random,GalerneGM11,xia2012compact,xia2014synthesizing,RaadDM16,GalerneL17siamis,GalerneLM17} is a simple texture model, synthesizing Gaussian textures is easy and requires few computational resources, however, it is not good at handling textures with complex geometry patterns. Gatys \etal~\cite{gatys2016image} used a CNN for texture synthesis, their method achieved good performance over a large scope of nature textures, but it failed when synthesizing textures with non-local structures and sometimes suffered from degraded quality~\cite{li2017diversified}. Liu \etal~\cite{liu2016texture} added extra spectrum constrain to textures, which improved the model's ability to model non-local structures. Sendik \etal \cite{sendik2017deep} further proposed using deep correlation matrix to model non-local structures and achieve better result than Liu \etal's ~\cite{liu2016texture}. Li \etal \cite{li2017diversified} proposed to use centred Gram matrix instead of Gram matrix to improve the quality of outputs.

The algorithm of neural style transfer is developed based on neural textures synthesis. The goal of style transfer is to transfer the ``style'' of one image to another, while keeping the ``content'' fixed. Although Gatys' vanilla style transfer algorithm~\cite{gatys2016image} can produce high quality stylish photos, it relies on expensive optimized based algorithm. To accelerate this time comsuming procedure, Johnson \etal~\cite{johnson2016perceptual} proposed a perceptual loss functions and a transformation network, this network can generate textures and stylish photos in a forward manner, hundreds of times faster than Gatys' vanilla algorithm. Ulyanov \etal~\cite{ulyanov2017improved} proposed to use instance normalization to improve the quality of outputs. To learn styles in one network, Li \etal~\cite{li2017diversified} and Dumoulin \etal~\cite{dumoulin2016learned} proposed new network structures.

In the past decades, tremendous studies have devoted to texture mixing. It aims at generating a ``averaged" texture from several textures.  Ruiters \etal~\cite{ruiters2010patch} presented a technique to interpolate between two textures based on patch-based method. Soheil \etal~\cite{darabi2012image} reported the state-of-the-art results of texture mixing via a image melding algorithm. In terms of statistic parametric models, texture mixing naturally means averaging statistics from different exemplars.
Peyre~\cite{peyre2010texture} proposed to use ``grouplet'' for synthesizing and mixing locally parallel textures. Joseph \etal~\cite{bar2001texture} proposed to use wavelet and a tree structure to model and mix textures.
Rabin \etal~\cite{rabin2011wasserstein} used optimal transport and pyramid decomposition to mix textures, the high dimensional Wasserstein metric is approximated by sliced one-dimensional Wasserstein metric. Optimal transport of Gaussian distribution has been studied to mix Gaussian textures~\cite{xia2014synthesizing}. These optimal-transport-based algorithm can generate homogeneous mixed textures, but they can not handle structured textures.

It is worth noticing that not all texture models are able to mix textures. e.g. in the prominent work of Portilla and Simoncelli's~\cite{portilla2000parametric}, they mentioned that the parametric space of their model is not convex, because linear interpolation of statistics will result in poor quality (patchwise) results. Even the state-of-the-art texture model proposed by Gatys \etal~\cite{gatys2015texture} suffered from the similar problem, as linear interpolation of Gram matrices only results in similar pathwise mixture~\cite{li2017diversified}.

\section{Problem Formulations}
\label{problem}
Denote $I \in \mathds{R} ^{\Omega \times d}$ as an image with $d$ channels defined on the grid $\Omega = \{0,...,M-1\} \times \{0,...N-1\}$. In particular, $d=1$ for grey-scale images and $d=3$ for color images.
For each pixel $p \in \Omega$, the value $I(p)$ is a $d$-dimensional vector, and for each channel $c \in \{0,..,d-1\}$ at location $p \in \Omega$, the value $I(p, c)$ is a real scalar.

\paragraph{Exemplar-based texture synthesis with CNNs.} Given a texture exemplar $I_{exp}$, the aim of exemplar-based texture synthesis is to produce new texture samples $I_{syn}$ that are as similar as possible to $I_{exp}$ regarding certain visual/perceptual measurements~\cite{portilla2000parametric}.
For instance, Zhu \etal~\cite{zhu1998filters} argued that $I_{syn}$ and $I_{ex}$ are equivalent on statistical feature sets,
$$
\{\mathbf{F}_{syn}^{(\ell_1)}, \ldots, \mathbf{F}_{syn}^{(\ell_k)}\} \sim \{\mathbf{F}_{exp}^{(\ell_1)}, \ldots, \mathbf{F}_{exp}^{(\ell_k)}\},
$$
where $\mathbf{F}_{\times} := \{\mathbf{F}_{\times}^{(\ell_1)}, \ldots, \mathbf{F}_{\times}^{(\ell_k)}\} = \mathscr{F} \circ I_{\times}$ are the sets of texture features extracted from $I_{\times}$ by a texture model $\mathscr{F}$, these models can be filter banks~\cite{zhu1998filters}, wavelets~\cite{portilla2000parametric} or Markovian models~\cite{efros1999texture}. The image $I_{syn}$ can thus be generated by feature projection~\cite{zhu1998filters,lu2016learning}. A survey of exemplar-based texture synthesis was recently provided in~\cite{RaadDDM17}.

In this paper, we are interested in exemplar-based texture model using deep CNN features~\cite{gatys2016image,sendik2017deep,gatys2015texture}, because of their capability to synthesize textures with complex structures.
This type of methods utilized a pre-learned deep CNN, $\mathscr{F}_{\textrm{CNN}}$, for texture description and suggested that $I_{syn}$ can be generated by matching deep features under Gram matrix or correlation matrix based similarity.
More precisely, one can initialize $I_{syn}$ with a random noise and pursue an optimal output by minimizing the following objective:
\begin{align}
\label{bojective_syn}
 \sum_{\ell=\ell_1}^{\ell_k} \Vert \mathcal{G}(\mathbf{F}_{syn}^{(\ell)}) -\mathcal{G}(\mathbf{F}_{exp}^{(\ell)}) \Vert_F^2,
\end{align}
where $\mathcal{G}(a)$ is the Gram measure of matrix $a$ and $\Vert \cdot \Vert_F$ denotes the Frobenius norm. The minimization problem in Eqn.~\eqref{bojective_syn} can be solved using back-propagation~\cite{gatys2015texture}.

\paragraph{Exemplar-based texture mixing with CNNs.}
Given two input texture exemplars $I_{exp_0}$ and $I_{exp_1}$, exemplar-based texture mixing aims to generate new textures whose visual and perceptual properties are drawn from both the inputs.
Denoting the deep features of the two inputs as $\mathbf{F}_{exp_0} = \mathscr{F}_{\textrm{CNN}} \circ I_{exp_0}$ and $\mathbf{F}_{exp_1} = \mathscr{F}_{\textrm{CNN}} \circ I_{exp_1}$ respectively, the mixing of $I_{exp_0}$ and $I_{exp_1}$ with ratio $\rho, \in [0,1]$ is to obtain $I_{syn}$, such that
$$
\mathbf{F}_{syn} \sim \{\rho \mathbf{F}_{exp_0},\, (1-\rho) \mathbf{F}_{exp_1}\},
$$
where $\mathbf{F}_{syn} = \mathscr{F}_{\textrm{CNN}} \circ I_{syn}$.
A straightforward solution is to pursue $I_{syn}$ by minimizing
\begin{align}
\label{LIA}
 \sum_{\ell=\ell_1}^{\ell_k} \Vert \rho \, \mathcal{G}(\mathbf{F}_{exp_0}^{(\ell)}) + (1-\rho)\, \mathcal{G}(\mathbf{F}_{exp_1}^{(\ell)}) - \mathcal{G}(\mathbf{F}_{syn}^{(\ell)}) \Vert_F^2,
\end{align}
which actually finds an $I_{syn}$ with linear interpolation of the Gram measurements. As we shall discuss in Section~\ref{experiments}, this mixing often produces results with conspicuous artifacts.

In what follows, we will develop a better means to interpolate the deep CNN features for mixing textures.

\section{Deep Texture Mixing with Gaussian Models}
\label{proposed_method}
Pioneered by Gatys \etal~\cite{gatys2015texture}, several studies have addressed texture synthesis with deep CNNs, e.g.~\cite{sendik2017deep,liu2016texture,li2017diversified,ulyanov2016texture}.
In this section, we first reveal that the works using Gram matrix~\cite{gatys2015texture}, centered Gram matrix~\cite{li2017diversified}, correlation~\cite{sendik2017deep} and spectrum~\cite{liu2016texture} can be unified and derived from a stationary Gaussian model.
We then show that this unified scheme enables us to mix textures by interpolating deep features generated from CNNs through a simple procedure.

\subsection{Gaussian Scheme for Deep Texture Synthesis}
\label{Gaussian_mixing}
Given $\mathbf{F} \in \mathds{R}^{U \times k}$, $U = Q \times M$, a feature map extracted by CNNs, deep texture synthesis methods~\cite{gatys2015texture,li2017diversified,sendik2017deep,liu2016texture} often first compute a statistic of $\mathbf{F}$, such as Gram matrix~\cite{gatys2015texture}, correlation~\cite{sendik2017deep}, centred Gram matrix~\cite{li2017diversified}, and spectrum~\cite{liu2016texture}.

\begin{itemize}
\item[-] {\bf Gram matrix $\mathcal{G}$.}
The Gram matrix $\mathcal{G}$ of $\mathbf{F} \in \mathds{R}^{U \times k}$ is defined as:
\begin{equation}
\label{G}
\mathcal{G}(i,j) = \frac{1}{|U|}\sum_{p \in U} \mathbf{F}(p,i) \mathbf{F}(p,j) ,\,\, 1 \leq i, j \leq k.\\
\end{equation}
$\mathcal{G}$ is a $k \times k$ positive semi-defined matrix.

\item[-] {\bf Centred Gram matrix $\bar{\mathcal{G}}$.}
Instead of using Gram matrix $\mathcal{G}$ of $\mathbf{F}$,  Li \etal~\cite{li2017diversified} suggested using centred Gram matrix $\bar{\mathcal{G}}$ to generate textures and reported that this approach resulted in better, or at least comparable, perceptual quality.
\begin{equation}
\bar{\mathcal{G}}(i,j) = \frac{1}{|U|}\sum_{p \in U}\Big( \mathbf{F}(p,i)-\mathbf{m}_i \Big) \Big(\mathbf{F}(p,j)-\mathbf{m}_j \Big),
\end{equation}
where $1 \leq i, j \leq k$, $\mathbf{m} \in \mathds{R}^{k}$ is the mean vector of $\mathbf{F} \in \mathds{R}^{U \times k}$,
\begin{align}
\label{mean}
\mathbf{m} = \frac{1}{|U|} \sum_{p \in U} \mathbf{F}(p).
\end{align}

\item[-] {\bf Correlation $\mathcal{S'}$.}
Recently, Sendik \etal~\cite{sendik2017deep} proposed using correlation $\mathcal{S}$ of $\mathbf{F}$ to synthesize textures, and reported the state-of-the-art results on non-local textures. Their deep correlations $\mathcal{S} \in \mathds{R}^{U \times k}$ of feature maps $\mathbf{F}$ are defined as

\begin{equation}
\label{Def_Cor_Pre}
\mathcal{S'}(p,n)= \sum_{p' \in U} w(p) \mathbf{F}(p',n)\mathbf{F}(p+p',n),
\end{equation}
in which $p=(i,j)$ is the offset vector, and $i \in [-Q/2, Q/2]$ and $j \in [-M/2, M/2]$, $w$ is the relative weight, defined by
\begin{equation}
\label{relative weight}
w(i,j) = ((Q-|i|)(M-|j|))^{-1}
\end{equation}

\item[-] {\bf Modified correlation $\mathcal{S}$.}
As we can see, Correlation $\mathcal{S'}$ relations ignores the relations between pixels whose horizontal distance are further than $Q/2$, or vertical distance further than $M/2$, we modify the correlation matrix by adding periodic boundary condition:
\begin{equation}
\label{Def_Cor}
\mathcal{S}(p,n)= \frac{1}{|U|}\sum_{p' \in U}\mathbf{F}(p',n)\mathbf{F}(p+p',n),
\end{equation}
where $1 \leq n \leq k, \, p \in U$. $\mathcal{S}(p)$ is a vector of length $k$.

Notice that these two definition is mathematically equivalent when $\mathbf{F}$ already satisfy periodic boundary condition. In later part of our paper, only the modified correlation matrix $\mathcal{S}$ will be used. As we will see in the experiment section, these two correlation matrix produce similar results.

\item[-] {\bf Spectrum $\mathcal{F}$.}
The Fourier spectrum $\mathcal{F}$ of $\mathbf{F}$ has been integrated into the process of texture synthesis~\cite{liu2016texture}.
$$
\mathcal{F} = \vert \hat{\mathbf{F}} \vert,
$$
where $\hat{a}$ denotes the Fourier transformation of $a$.
\end{itemize}

\paragraph{Stationary Gaussian model}
For feature maps $\mathbf{F} \in \mathds{R}^{U \times k}$, the associated stationary Gaussian model $\mu(\mathbf{m}, \mathcal{C})$ can be obtained by estimating mean $\mathbf{m} \in \mathds{R}^{k}$ using Eqn.~\eqref{mean} and covariance $\mathcal{C} \in \mathds{R}^{U \times k \times k}$ using  the following equation~\cite{xia2014synthesizing}:
\begin{align}
\label{covariance}
\mathcal{C}(p,i,j) = \frac{1}{|U|}\sum_{p' \in U} \Big(\mathbf{F}(p', i)-\mathbf{m}_i \Big) \Big(\mathbf{F}(p+p',j)-\mathbf{m}_j \Big),
\end{align}
where $p \in U , \,\,1 \leq i,\,\, j \leq k$, and $\mathcal{C}(p)$ is a matrix of size $k \times k$.

The following proposition provides the connections between the above four measurements and the stationary Gaussian model $\mu(\mathbf{m}, \mathcal{C})$.
\begin{proposition}
\label{proposition_gaussian}
Given feature maps $\mathbf{F} \in \mathds{R}^{U \times k}$, its Gram matrix $\mathcal{G}$, centred Gram matrix $\bar{\mathcal{G}}$, correlation $\mathcal{S}$ and spectrum $\mathcal{F}$ can be derived from a stationary Gaussian model:
\begin{align}
\label{Gram}
\mathcal{G} &=\mathcal{C}(0) +  \mathbf{m}\mathbf{m}^T,\\
\label{center_Gram}
\bar{\mathcal{G}} &=\mathcal{C}(0), \\
\label{correlation}
\forall p \in U,\,\, \mathcal{S}(p) &= diag(\mathcal{C}(p)) + \mathbf{m} \odot \mathbf{m}, \\
\label{spectrum}
\forall \omega \in U,\,\,\mathcal{F}(\omega) &= (|U| \vert \hat{\mathcal{S}}(\omega) \vert)^{\frac{1}{2}}.
\end{align}
where $^T$ is the transpose operator, $\odot$ denotes the componentwise product, and $\hat{\cdot}$ is the Fourier transformation.
\end{proposition}

The derivations of Eqn.~\eqref{Gram}~\eqref{center_Gram}~\eqref{correlation} are straightforward.
Eqn.~\eqref{spectrum} occurs because the correlation $\mathcal{S}$ is the auto-correlation of the feature map $\mathbf{F}$, which offers
$$
  \hat{\mathcal{S}}(\omega) = \frac{1}{|U|} \hat{\mathbf{F}}(\omega) \odot \hat{\mathbf{F}}(\omega)^*,
$$
 with $*$ denoting the conjugate transpose.

It is worth noticing that the Proposition~\ref{proposition_gaussian} can explain the experimental observations reported  by~\cite{gatys2015texture,sendik2017deep,liu2016texture,li2017diversified}: using Gram matrix and central Gram matrix generate comparable synthesized results, while using spectrum~\cite{liu2016texture} and correlation~\cite{sendik2017deep} are both effective in synthesizing non-local structures.

\subsection{Interpolating Deep Statistics via Gaussian Model}
In Gatys' texture model~\cite{gatys2015texture}, mixing textures corresponds to interpolating the Gram matrices. Formally, the goal is to pursue a continuous function $\mathcal{G}(\rho)$, $\rho \in [0,1]$, such that $\mathcal{G}(\rho) = \mathcal{G}_\rho$ when $\rho=0, 1$.
Obviously, the solution is not unique. For instance, a simple linear interpolation satisfies this requirement, however, linear interpolations of Gram matrices does not necessarily result in Gram matrices, and it performs poorly for texture mixing (see experiment in Fig.~\ref{Compare_All} and \ref{CNN}), primarily because it ignores the manifold of Gram matrix $\mathcal{G}$.

Given two Gaussian models $\mu_0 = (\mathbf{m}_0, \mathcal{C}_0)$ and  $\mu_1 = (\mathbf{m}_1, \mathcal{C}_1)$, the problem of interpolating Gaussian models can be stated as: find a continuous function $\mu({\rho}) = (\mathbf{m}_{\rho}, \mathcal{C}_{\rho})$, ${\rho} \in [0,1]$ such that $\mu({\rho}) = \mu_{\rho}$ when ${\rho}=0, 1$.

The significance of roposition~\ref{proposition_gaussian} is that it reduce the problem of interpolation of Gram matrices $\mathcal{G}$ to the problem of interpolation of Gaussian models $\mu$. Formally, Proposition~\ref{proposition_gaussian} demonstrates that, given feature maps $\mathbf{F}$, the Gram matrix $\mathcal{G}$ and Gaussian model $\mu$, there exists a continuous function $f_{\mathcal{G}}$, such that
\begin{equation}
\mathcal{G}=f_{\mathcal{G}}(\mu).
\end{equation}
Therefore,  given feature maps $\mathbf{F}_{0}$ and $\mathbf{F}_{1}$, their Gram matrices $\mathcal{G}_0$ and $\mathcal{G}_1$, Gaussian models $\mu_0$ and $\mu_1$ respectively, if we can find a continuous function $\mu({\rho}) = (\mathbf{m}_{\rho}, \mathcal{C}_{\rho})$, ${\rho} \in [0,1]$ such that $\mu({\rho}) = \mu_{\rho}$ when ${\rho}=0, 1$. We can consequently obtain a continuous function as a composition of  $\mu({\rho})$ and $f_{\mathcal{G}}$:
\begin{equation}
\label{composition}
\mathcal{G}(\rho)=f_{\mathcal{G}}(\mu(\rho)),
\end{equation}
where $\mathcal{G}(0)=\mathcal{G}_0$ and $\mathcal{G}(1)=\mathcal{G}_1$, so function $\mathcal{G}(\rho)$ in Eqn.~\eqref{composition} can be used to interpolate between Gram matrices $\mathcal{G}_0$ and $\mathcal{G}_1$.



Possible solutions to the problem of interpolation of Gaussian models include linear interpolation, Fisher-Rao interpolation~\cite{atkinson1981rao} and optimal transport interpolation~\cite{xia2014synthesizing}, but linear interpolated $\mu({\rho})$ is no longer Gaussian, and no explicit formula is know for high dimensional Fisher-Rao interpolation. Alternatively, optimal transport interpolation provide a closed-form solution to the problem and it can be shown that interpolated $\mu({\rho})$ remains Gaussian~\cite{xia2014synthesizing}.

According to~\cite{xia2014synthesizing}, given feature maps $\mathbf{F}_{0}$ and $\mathbf{F}_{1}$ whose Gaussian models are $\mu_0$ and $\mu_1$ respectively, the feature maps $\mathbf{F}_t$ of interpolated $\mu_t$ can be calculated by using the following formula in Fourier domain,
\begin{align}
\label{geodesic0}
&\forall {\rho} \in [0,1],\;\hat{\mathbf{F}}_{\rho} =(1-{\rho})\hat{\mathbf{F}}_0 + {\rho}\hat{\mathbf{G}},\\
\label{geodesic1}
&\forall w,\; \hat{\mathbf{G}}(w)=\hat{\mathbf{F}}_1(w)\frac{\hat{\mathbf{F}}_1(w)^*\hat{\mathbf{F}}_0(w)}{|\hat{\mathbf{F}}_1(w)^*\hat{\mathbf{F}}_0(w)|},
\end{align}
where $\hat{a}$ denotes the Fourier transformation and $a^*$ is the conjugate transpose of $a$.

Note that practical algorithm for mixing Gram matrix or correlation does not even requires computing Gaussian model, because $\mathcal{G}_{\rho}$ and $\mathcal{S}_{\rho}$ can be derived directly from the feature map $\mathbf{F}_{\rho}$ using Eqn.~\eqref{G} and \eqref{Def_Cor}. Combining formula Eqn.~\eqref{geodesic0}, \eqref{geodesic1}, ~\eqref{Def_Cor} and \eqref{G}, we obtain the following proposition.

\begin{proposition}
\label{Summary}
Given feature maps $\mathbf{F}_0$, $\mathbf{F}_1$ $\in \mathds{R}^{U \times k}$, and a relative weight $\rho \in [0,1]$, the interpolated Gram matrix and correlation matrix can be written as follows:
\begin{align}
&\forall \rho \in [0,1], \; \hat{\mathbf{F}}_{\rho} = (1-\rho)\hat{\mathbf{F}}_0 + {\rho} \hat{\mathbf{G}}, \\
&\forall w,\; \hat{\mathbf{G}}(w)=\hat{\mathbf{F}}_1(w) \frac{\hat{\mathbf{F}}_1(w)^*\hat{\mathbf{F}}_0(w)}{|\hat{\mathbf{F}}_1(w)^*\hat{\mathbf{F}}_0(w)|} ,\\
\label{Summary_G}
&\mathcal{G}_{\rho} (i,j) = \frac{1}{|U|} \sum_{p \in U}\mathbf{F}_{\rho} (p,i)\mathbf{F}_{\rho} (p,j),\\
\label{Summary_S}
&\mathcal{S}_{\rho} (p,g) = \frac{1}{|U|}\sum_{p' \in U}\mathbf{F}_{\rho} (p',g)\mathbf{F}_{\rho} (p+p',g),
\end{align}
where $p \in U , \,\,1 \leq i,\, j,\,g \leq k$.
\end{proposition}

Proposition~\ref{Summary} provides the closed-form computation of the mixed Gram matrix and correlation matrix using optimal transport. All the computations are implemented on feature maps $\mathbf{F}$.
A conceptual illustration is provided in Fig.~\ref{Flow}.

\section{Implementation Details}
\label{implementations}
This section presents the implementation details of our Gaussian scheme for texture mixing.
For simplicity, we follow the texture synthesis pipeline proposed by Gatys~\etal~\cite{gatys2015texture}. However, note that our method can also be used to interpolate centred Gram matrix, correlation matrix and spectrum.
We also apply our scheme to styles morphing.

\subsection{Texture Mixing}
\label{texture_mixing}
As presented in Section~\ref{Gaussian_mixing}, interpolating Gram matrices $\mathcal{G}_{exp_0}$ and $\mathcal{G}_{exp_1}$ can be computed by mixing its feature maps $\mathbf{F}_{exp_0}$ and $\mathbf{F}_{exp_1}$. Therefore, our Gaussian scheme for exemplar-based texture mixing with CNNs can be summarized by algorithm~\ref{algorithm}. For simplicity, we use only two styles/textures in our algorithm, but it can be extended to morph more styles/textures without difficulty.

\begin{algorithm}[ht!]
	\caption{Deep texture mixing with Gaussian}\label{algorithm}
	\begin{algorithmic}
	\Require exemplar textures $I_{exp_0}$, $I_{exp_1}$, $\rho \in [0,1]$, a pre-trained CNN texture descriptor $\mathscr{F}_{\textrm{CNN}}$ .
	\Ensure a sample of mixed texture $I_{syn}$.
        \For{$i= 0, 1$}
        \State  $\big\{ \mathbf{F}_i^{(\ell_1)}, \mathbf{F}_i^{(\ell_2)}\ldots,  \mathbf{F}_i^{(\ell_k)}\big\}\, \leftarrow \, \mathscr{F}_{\textrm{CNN}} \circ I_{exp_i}$.
        \EndFor
        \For{$\ell= \{\ell_1, \ell_2, \ldots, \ell_k\}$}
        \State  $\mathbf{F}_\rho^{(\ell)} \, \leftarrow \, Mixing(\mathbf{F}_0^{(\ell)}, \mathbf{F}_1^{(\ell)}, \rho)$ using Eqn.~\eqref{geodesic0}~\eqref{geodesic1};
        \State compute $\mathcal{G}_{\rho}^{(\ell)}$ or $\mathcal{S}_{\rho}^{(\ell)}$ from $\mathbf{F}_\rho^{(\ell)}$ using Eqn.~\eqref{Summary_G}~\eqref{Summary_S}.
        \EndFor
        \State generate $I_{syn}$ by minimizing Eqn~\eqref{Gen texture}.
	\end{algorithmic}
\end{algorithm}

More precisely, two input exemplars, $I_{exp_0}$ and $I_{exp_1}$,  are firstly feed to a pre-trained deep CNN (\ie $\mathscr{F}_{\textrm{CNN}}$) respectively. The outputs of selected layers $\{\ell_1, \ell_2, \ldots, \ell_k\}$ are then extracted as $\{ \mathbf{F}_i^{(\ell_1)}, \mathbf{F}_i^{(\ell_2)}\ldots,  \mathbf{F}_i^{(\ell_k)}\}$, with $i=0,1$. A Gram matrix $\mathcal{G}_i^{(\ell)}$ is computed for each $\mathbf{F}_i^{(\ell)}$.
In order to interpolate between $\{ \mathcal{G}_i^{(\ell)}\}_{0,1}$ with a relative weight $\rho \in [0,1]$, we only need to implicitly interpolate $ \{\mathbf{F}_i^{(\ell)}\}_{0,1}$ using Equation~\eqref{geodesic0}~\eqref{geodesic1}.
The interpolated Gram matrix $\mathcal{G}_{\rho}^{(\ell)}$ is then computed from the mixed feature maps $\mathbf{F}_{\rho}^{(\ell)}$ by Equation~\eqref{Summary_G}.
To generate new textures, as in~\cite{gatys2015texture}, back-propagation is used to generate an image $I_{syn}$ whose Gram matrices $\{ \mathcal{G}_{syn}^{(\ell_1)}, \ldots, \mathcal{G}_{syn}^{(\ell_k)} \}$ are as similar as possible to those of the exemplar $\{ \mathcal{G}_{\rho}^{(\ell_1)}, \ldots, \mathcal{G}_{\rho}^{(\ell_k)} \}$, \ie by minimizing:
\begin{equation}
\label{Gen texture}
         \sum_{\ell= \ell_1}^{\ell_k} \Vert \mathcal{G}_{\rho}^{(\ell)} - \mathcal{G}_{syn}^{(\ell)} \Vert^2_F.
\end{equation}

\subsection{Styles Morphing}
Our scheme can also be applied to morphing the styles of two images.
Given an original image $I_{ori}$ and two style images $I_{sty_0}$ and $I_{sty_1}$, the goal of {\em styles morphing} is to interpolate between the style $I_{sty_0}$ and style $I_{sty_1}$, while keeping the content of $I_{ori}$.

Similar to the texture model, a Gram matrix is used to parametrized the style.
Denote $\mathbf{F}_{ori}$, $\mathbf{F}_{syn}$ as the feature maps of $I_{ori}$ and synthesized image. Let $\mathcal{G}_{sty_0}$, $\mathcal{G}_{sty_1}$ and $\mathcal{G}_{syn}$ be the Gram matrix of the style images and the synthesized image.
Such an image $I_{syn}$ can be generated by minimizing
\begin{equation}
\sum_{\ell = \ell_1}^{\ell_k} \Vert \mathcal{G}_{\rho}^{(\ell)} - \mathcal{G}_{syn}^{(\ell)} \Vert_F^2 + \alpha \Vert \mathbf{F}_{ori}^{(\ell)} - \mathbf{F}_{syn}^{(\ell)} \Vert_F^2
\end{equation}
where $\mathcal{G}_{\rho}^{(\ell)}$ is an interpolated Gram matrix of $\mathcal{G}_{sty_0}$ and $\mathcal{G}_{sty_1}$ with a weight $\rho \in [0,1]$.
$\alpha$ is a parameter to control the degree of style bending.

Similar to our texture mixing algorithm described in section~\ref{texture_mixing}, $\mathcal{G}_{\rho}^{(\ell)}$ can be computed with Eqns.~\eqref{Summary_G}, after interpolating $\mathbf{F}_{sty_0}$ and $\mathbf{F}_{sty_1}$ using Equation~\eqref{geodesic0}~\eqref{geodesic1}.

Note that, because our mixing process is in closed form, it costs almost the same time as Gatys' texture synthesizing algorithm~\cite{gatys2016image}.

\section{Experimental Results and Analysis}
\label{experiments}
In this section, we firstly evaluated the performance of our modified correlation matrix $\mathcal{S}$ on non-local textures, and compared them with Sendik's result~\cite{sendik2017deep}, where we showed that our modified correlation matrix can produce comparable results with theirs. Then we showed our texture mixing scheme is effective in both non-periodic textures and periodic textures, and compared the results with other sate-of-the-art texture mixing algorithms. Finally, we applied our algorithm to style transfer, where we are able to produce high visual quality stylish photo with ``intermediate'' styles.

For all the experiments, we use VGG-19~\cite{simonyan2014very} network pre-trained on ImageNet dataset~\cite{imagenet_cvpr09}. In texture mixing and style morphing experiments, $10$ values of relative weight $\rho$ are used: $\rho=$ $\frac{0}{9}$, $\frac{1}{9}$, $\frac{2}{9}$, $\frac{3}{9}$,$\frac{4}{9}$, $\frac{5}{9}$, $\frac{6}{9}$, $\frac{7}{9}$, $\frac{8}{9}$, $\frac{9}{9}$. In texture mixing experiments, input images are down-sampled to $(256, 256)$ or (128,128) depending on its original size, and in style morphing experiments, input images are down-sampled to $(256, 256)$.
All experimental results are available at \url{http://captain.whu.edu.cn/TexMixDeepG}.

\subsection{Comparisons of Correlations Matrices}
In order to validate the efficiency of our proposition on the correlation matrix, given in Proposition~\ref{proposition_gaussian}, we compare the results using our modified correlation matrix $\mathcal{S}$ and Sendik's correlation matrix $\mathcal{S}'$~\cite{sendik2017deep} on texture synthesis. More precisely, we follow the same experimental settings as in~\cite{sendik2017deep}: using
layers \texttt{pool1}, \texttt{pool2}, \texttt{pool3}, \texttt{pool4} for Gram loss, using layer \texttt{pool2} for correlation loss, and using \texttt{conv1} for smooth loss. Moreover, each layer of the network is equally weighted. For experiments, each input image is re-scaled to $256 \times 256$ pixels, and each output image is initialized as white Gaussian noise. For optimization, the L-BFGS algorithm~\cite{Zhu1994L} is used.

\begin{figure*}[htb!]
	\begin{center}
	\TabHeight{
		\includegraphics[width=0.1\linewidth]{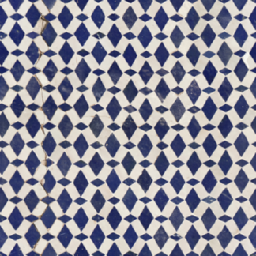}
		\includegraphics[width=0.1\linewidth]{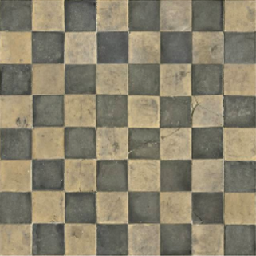}
		\includegraphics[width=0.1\linewidth]{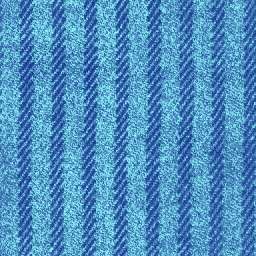}
		\includegraphics[width=0.1\linewidth]{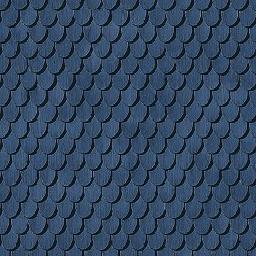}
		
		\includegraphics[width=0.1\linewidth]{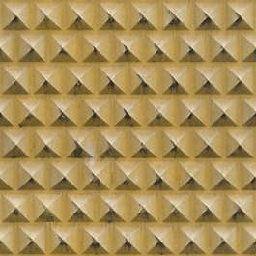}
		\includegraphics[width=0.1\linewidth]{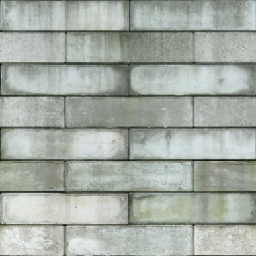}
		\includegraphics[width=0.1\linewidth]{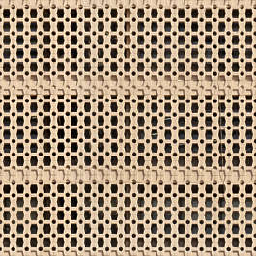}
		\includegraphics[width=0.1\linewidth]{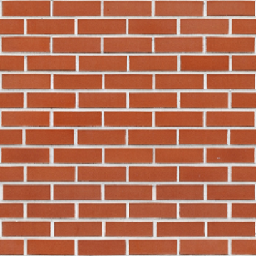}
		\includegraphics[width=0.1\linewidth]{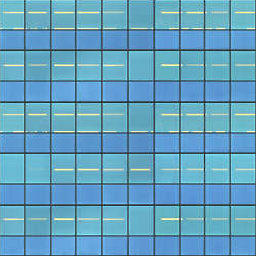}
	}\\
	\TabHeight{
		\includegraphics[width=0.1\linewidth]{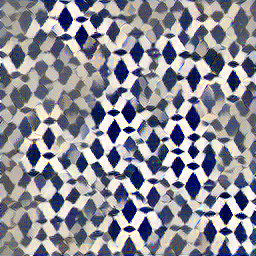}
		\includegraphics[width=0.1\linewidth]{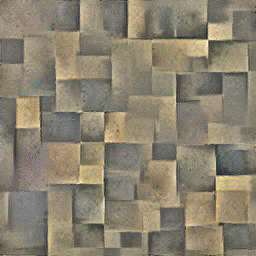}
		\includegraphics[width=0.1\linewidth]{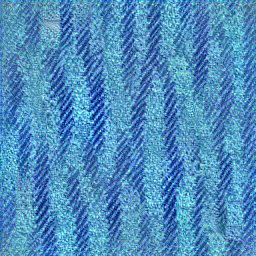}
		\includegraphics[width=0.1\linewidth]{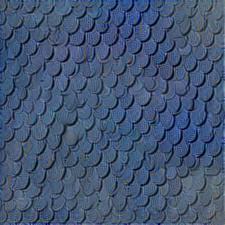}
		
		\includegraphics[width=0.1\linewidth]{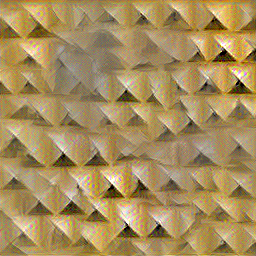}
		\includegraphics[width=0.1\linewidth]{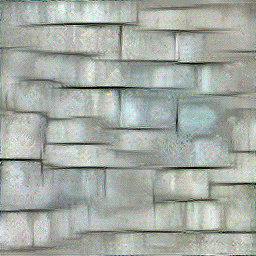}
		\includegraphics[width=0.1\linewidth]{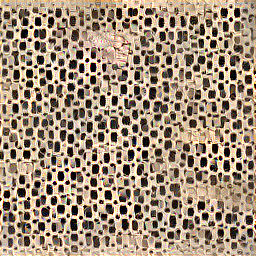}
		\includegraphics[width=0.1\linewidth]{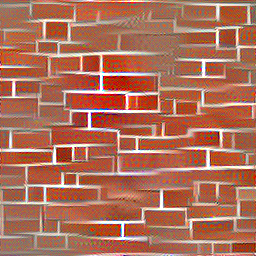}
		\includegraphics[width=0.1\linewidth]{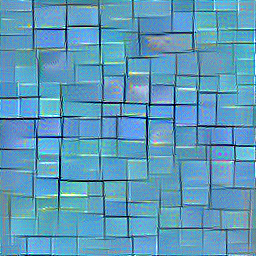}
	}\\
	\TabHeight{
		\includegraphics[width=0.1\linewidth]{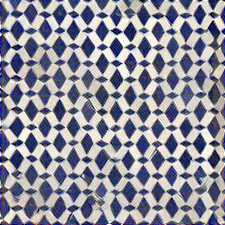}
		\includegraphics[width=0.1\linewidth]{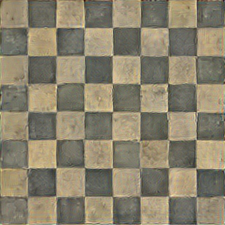}
		\includegraphics[width=0.1\linewidth]{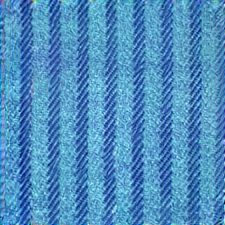}
		\includegraphics[width=0.1\linewidth]{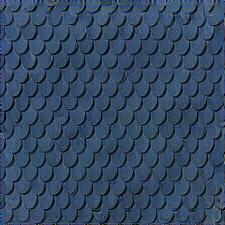}

		\includegraphics[width=0.1\linewidth]{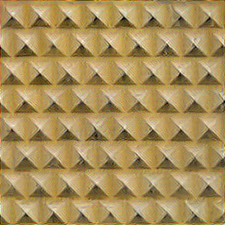}
		\includegraphics[width=0.1\linewidth]{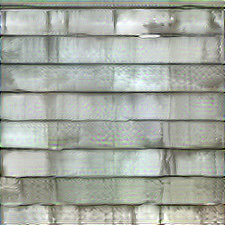}
		\includegraphics[width=0.1\linewidth]{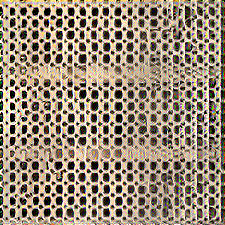}
		\includegraphics[width=0.1\linewidth]{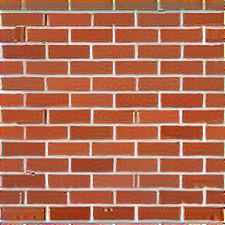}
		\includegraphics[width=0.1\linewidth]{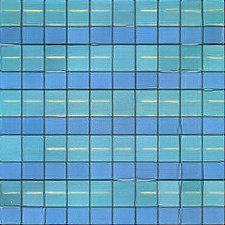}
	}\\
	\TabHeight{
		\includegraphics[width=0.1\linewidth]{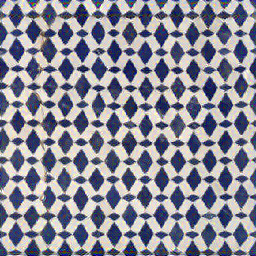}
		\includegraphics[width=0.1\linewidth]{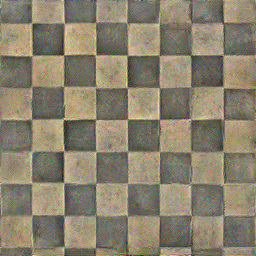}
		\includegraphics[width=0.1\linewidth]{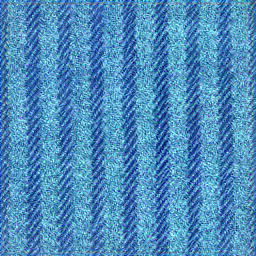}
		\includegraphics[width=0.1\linewidth]{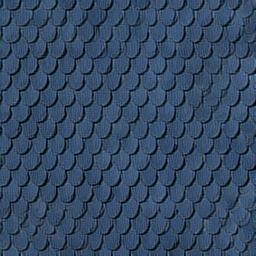}
		\includegraphics[width=0.1\linewidth]{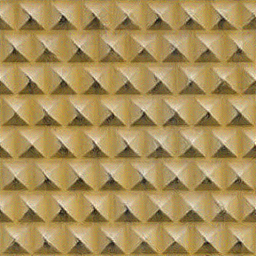}
		\includegraphics[width=0.1\linewidth]{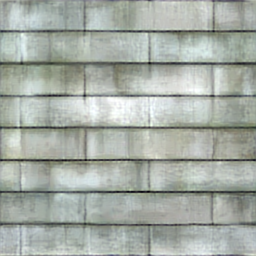}
		\includegraphics[width=0.1\linewidth]{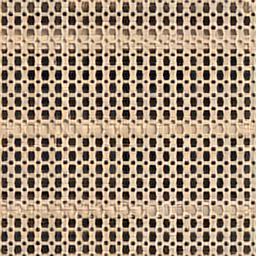}
		\includegraphics[width=0.1\linewidth]{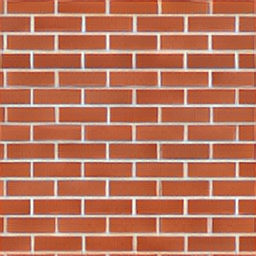}
		\includegraphics[width=0.1\linewidth]{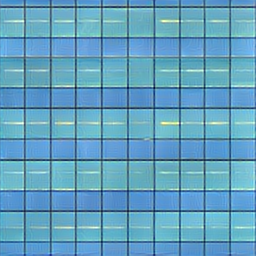}
	}\\

	\end{center}
\vspace{-3mm}
\caption{Comparison of using different correlation matrices.
From top to bottom, {\bf $1st$ row}: input images;  {\bf $2nd$ row}: results achieved using Gram matrix $\mathcal{G}$ as in Gatys~\cite{gatys2015texture}; {\bf $3rd$ row}: results achieved using the Sendik's correlation matrix $\mathcal{S}'$ in~\cite{sendik2017deep}; {\bf $4th$ row}:
results achieved using our modified correlation matrix $\mathcal{S}$ described in Eqn.~\eqref{correlation}.
Observe that using both $\mathcal{S}$ and $\mathcal{S}'$ can synthesize well non-local textures, but using Gram matrix can not.}
\label{Com_Cor}
\end{figure*}

In Fig~\ref{Com_Cor}, one can see that our modified correlation matrix $\mathcal{S}$ produced comparable results to that of Sendik's, while Gatys' Gram matrix failed to capture non-local structures.

\begin{figure*}[htb!]
	\begin{center}
		\includegraphics[width=0.95\linewidth]{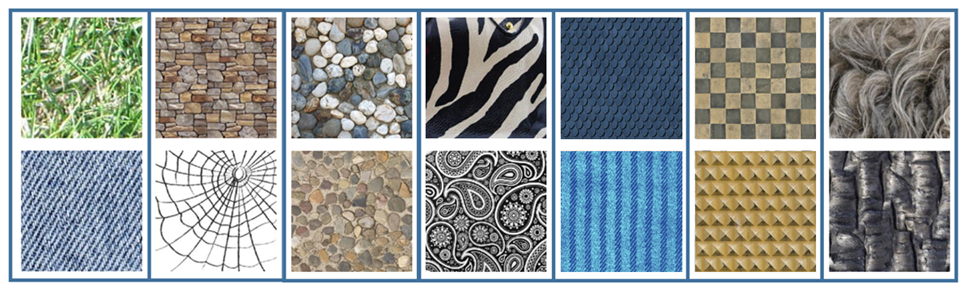}
	\end{center}
\vspace{-3mm}
\caption{Pairs of textures (in column) used for texture mixing experiments.}
\label{Textures}
\end{figure*}

\subsection{Comparisons with State-of-the-art Texture Mixing Methods}
This section evaluates our texture mixing method by comparing it with several texture mixing algorithms. We also compared our algorithm with TextureNet~\cite{ulyanov2017improved}, which enable us to generate mixed textures instantly. The compared mixing algorithms are listed as follows:

\begin{itemize}
\item[-] {\textbf{GaussTexton}~\cite{xia2014synthesizing}}:
A simple but fast texture mixing algorithm based on stationary Gaussian models.
\item[-] {\textbf{Image Melding}~\cite{darabi2012image}}:
An efficient texture mixing method based on patch match algorithm.
\item[-] {\textbf{Diversified Feed-forward Networks (DFN)}~\cite{li2017diversified}}:
Texture mixing by using linear combinations of different ``selectors''.
\item[-] {\textbf{Linear Interpolation Algorithm (LIA)}}:
Linear interpolation of Gram matrix as given in Eqn.~\eqref{LIA}.

\item[-] {\textbf{Our scheme + TextureNet}}:
The combination of Improved TextureNet~\cite{ulyanov2017improved} and our method, {\em i.e.} the Gram matrices used in TextureNet is mixed using our algorithm.

\item[-] {\textbf{LIA + TextureNet}}:
The combination of Improved TextureNet~\cite{ulyanov2017improved} and linear interpolation algorithm, {\em i.e.} the Gram matrices used in TextureNet is mixed using Eqn.~\eqref{LIA}.
\end{itemize}

Following the settings of Gatys~\cite{gatys2015texture}, \texttt{conv1\_1}, \texttt{pool1}, \texttt{pool2}, \texttt{pool3} and \texttt{pool4} layers are selected to exert constrains.  Pairs of texture exemplars used in the experiment are shown in Fig.~\ref{Textures}, which are taken from DTD dataset~\cite{cimpoi14describing} or collected from Internet.

\begin{figure*}[htb!]
	\begin{center}
	\TabHeight{
		\includegraphics[width=0.096\linewidth]{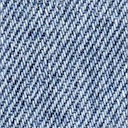}
		\includegraphics[width=0.096\linewidth]{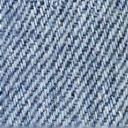}
		\includegraphics[width=0.096\linewidth]{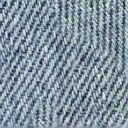}
		\includegraphics[width=0.096\linewidth]{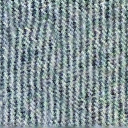}
		\includegraphics[width=0.096\linewidth]{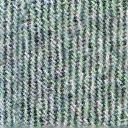}
		\includegraphics[width=0.096\linewidth]{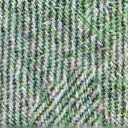}
		\includegraphics[width=0.096\linewidth]{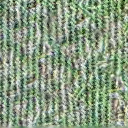}
		\includegraphics[width=0.096\linewidth]{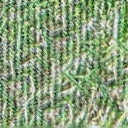}
		\includegraphics[width=0.096\linewidth]{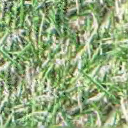}
		\includegraphics[width=0.096\linewidth]{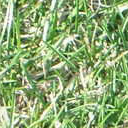}
	}\\
	\TabHeight{
		\includegraphics[width=0.096\linewidth]{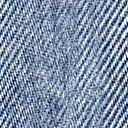}
		\includegraphics[width=0.096\linewidth]{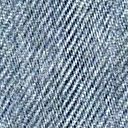}
		\includegraphics[width=0.096\linewidth]{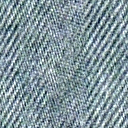}
		\includegraphics[width=0.096\linewidth]{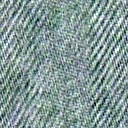}
		\includegraphics[width=0.096\linewidth]{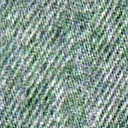}
		\includegraphics[width=0.096\linewidth]{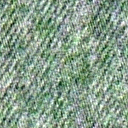}
		\includegraphics[width=0.096\linewidth]{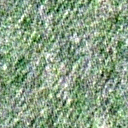}
		\includegraphics[width=0.096\linewidth]{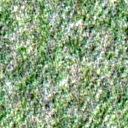}
		\includegraphics[width=0.096\linewidth]{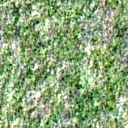}
		\includegraphics[width=0.096\linewidth]{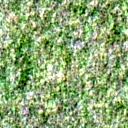}
	}\\
	\TabHeight{
		\includegraphics[width=0.096\linewidth]{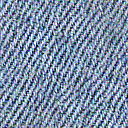}
		\includegraphics[width=0.096\linewidth]{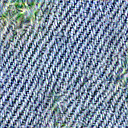}
		\includegraphics[width=0.096\linewidth]{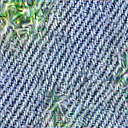}
		\includegraphics[width=0.096\linewidth]{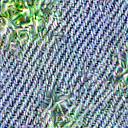}
		\includegraphics[width=0.096\linewidth]{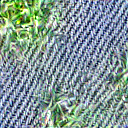}
		\includegraphics[width=0.096\linewidth]{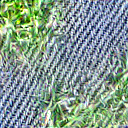}
		\includegraphics[width=0.096\linewidth]{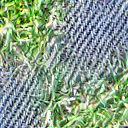}
		\includegraphics[width=0.096\linewidth]{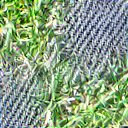}
		\includegraphics[width=0.096\linewidth]{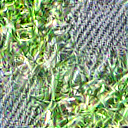}
		\includegraphics[width=0.096\linewidth]{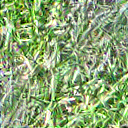}
	}\\
	\TabHeight{
		\includegraphics[width=0.096\linewidth]{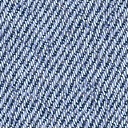}
		\includegraphics[width=0.096\linewidth]{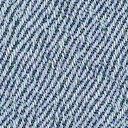}
		\includegraphics[width=0.096\linewidth]{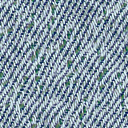}
		\includegraphics[width=0.096\linewidth]{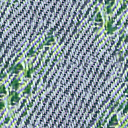}
		\includegraphics[width=0.096\linewidth]{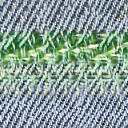}
		\includegraphics[width=0.096\linewidth]{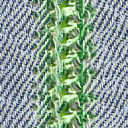}
		\includegraphics[width=0.096\linewidth]{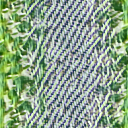}
		\includegraphics[width=0.096\linewidth]{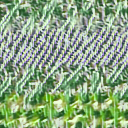}
		\includegraphics[width=0.096\linewidth]{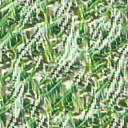}
		\includegraphics[width=0.096\linewidth]{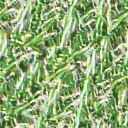}
	}\\
	\TabHeight{
		\includegraphics[width=0.096\linewidth]{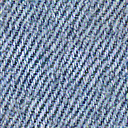}
		\includegraphics[width=0.096\linewidth]{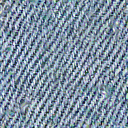}
		\includegraphics[width=0.096\linewidth]{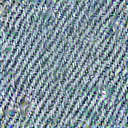}
		\includegraphics[width=0.096\linewidth]{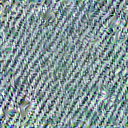}
		\includegraphics[width=0.096\linewidth]{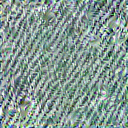}
		\includegraphics[width=0.096\linewidth]{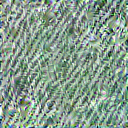}
		\includegraphics[width=0.096\linewidth]{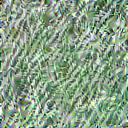}
		\includegraphics[width=0.096\linewidth]{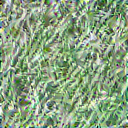}
		\includegraphics[width=0.096\linewidth]{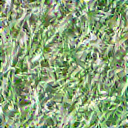}
		\includegraphics[width=0.096\linewidth]{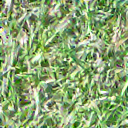}
	}\\	
	\TabHeight{
		\includegraphics[width=0.096\linewidth]{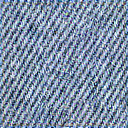}
		\includegraphics[width=0.096\linewidth]{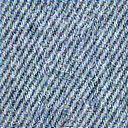}
		\includegraphics[width=0.096\linewidth]{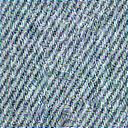}
		\includegraphics[width=0.096\linewidth]{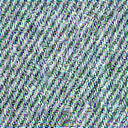}
		\includegraphics[width=0.096\linewidth]{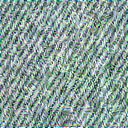}
		\includegraphics[width=0.096\linewidth]{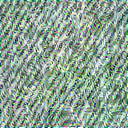}
		\includegraphics[width=0.096\linewidth]{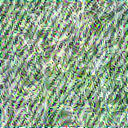}
		\includegraphics[width=0.096\linewidth]{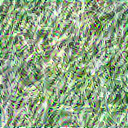}
		\includegraphics[width=0.096\linewidth]{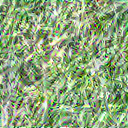}
		\includegraphics[width=0.096\linewidth]{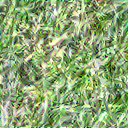}
	}\\
	\TabHeight{
		\includegraphics[width=0.096\linewidth]{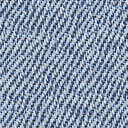}
		\includegraphics[width=0.096\linewidth]{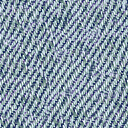}
		\includegraphics[width=0.096\linewidth]{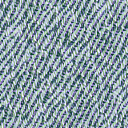}
		\includegraphics[width=0.096\linewidth]{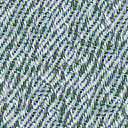}
		\includegraphics[width=0.096\linewidth]{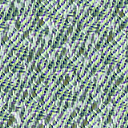}
		\includegraphics[width=0.096\linewidth]{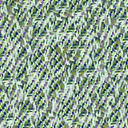}
		\includegraphics[width=0.096\linewidth]{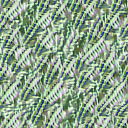}
		\includegraphics[width=0.096\linewidth]{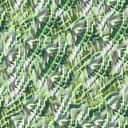}
		\includegraphics[width=0.096\linewidth]{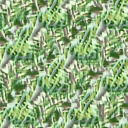}
		\includegraphics[width=0.096\linewidth]{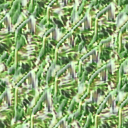}
	}\\
	\end{center}
\vspace{-3mm}
\caption{Experiments on mixing micro textures, by using Image Melding (1-st row)~\cite{darabi2012image},
using GaussTexton (2-nd row),
LIA (3-rd row),
LIA + TextureNet (4-th row).
our mixing scheme + Gram matrix (5-th row),
our mixing scheme + correlation (6-th row),
our mixing scheme + TextureNet (7-th row).
Observe that our scheme, both with or without TextureNet, can smoothly interpolate between the two inputs without producing new structures. See text for more details.}
\label{Compare_All}
\end{figure*}

\begin{figure*}[htb!]
	\begin{center}
	\TabHeight{
		\includegraphics[width=0.096\linewidth]{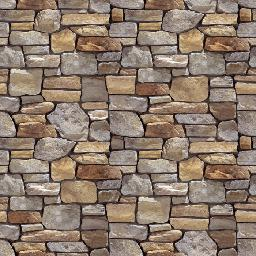}
		\includegraphics[width=0.096\linewidth]{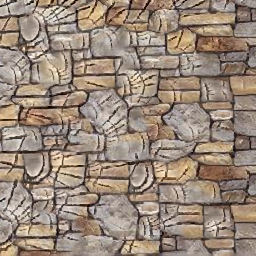}
		\includegraphics[width=0.096\linewidth]{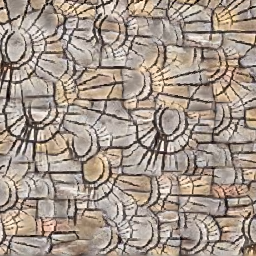}
		\includegraphics[width=0.096\linewidth]{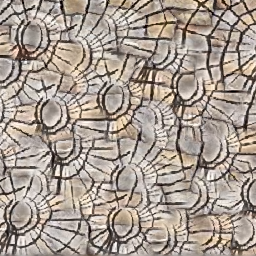}
		\includegraphics[width=0.096\linewidth]{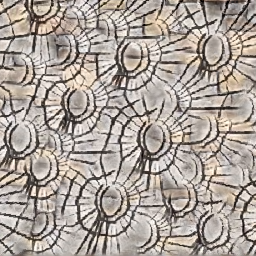}
		\includegraphics[width=0.096\linewidth]{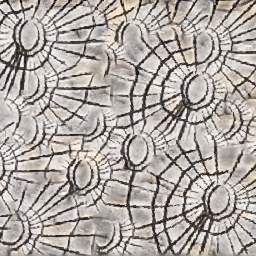}
		\includegraphics[width=0.096\linewidth]{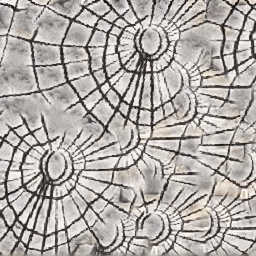}
		\includegraphics[width=0.096\linewidth]{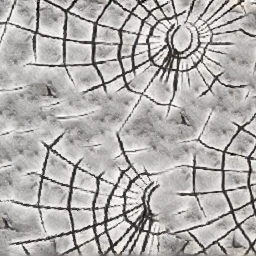}
		\includegraphics[width=0.096\linewidth]{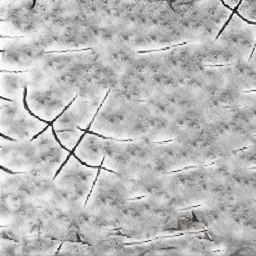}
		\includegraphics[width=0.096\linewidth]{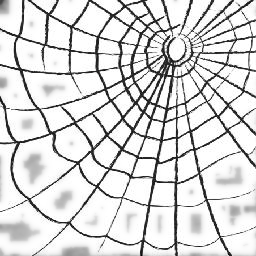}
	}\\
	\TabHeight{
		\includegraphics[width=0.096\linewidth]{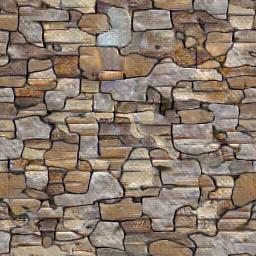}
		\includegraphics[width=0.096\linewidth]{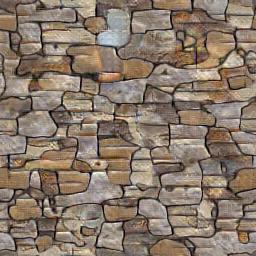}
		\includegraphics[width=0.096\linewidth]{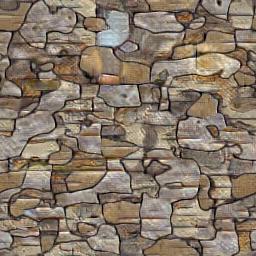}
		\includegraphics[width=0.096\linewidth]{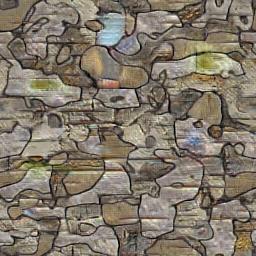}
		\includegraphics[width=0.096\linewidth]{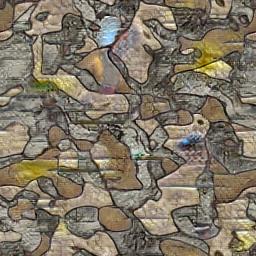}
		\includegraphics[width=0.096\linewidth]{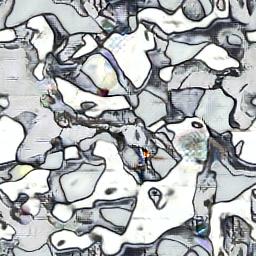}
		\includegraphics[width=0.096\linewidth]{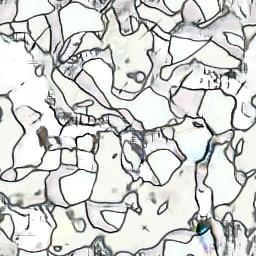}
		\includegraphics[width=0.096\linewidth]{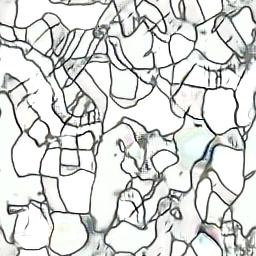}
		\includegraphics[width=0.096\linewidth]{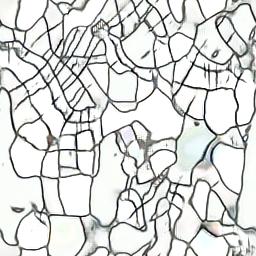}
		\includegraphics[width=0.096\linewidth]{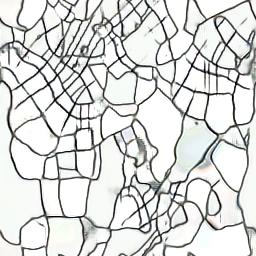}
	}\\
	\TabHeight{
		\includegraphics[width=0.096\linewidth]{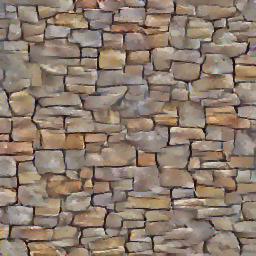}
		\includegraphics[width=0.096\linewidth]{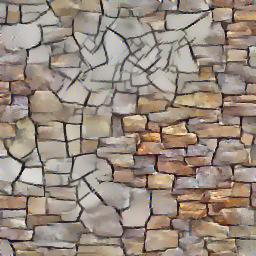}
		\includegraphics[width=0.096\linewidth]{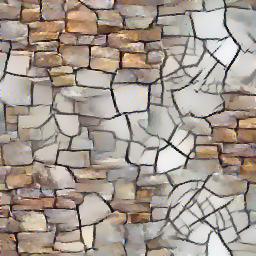}
		\includegraphics[width=0.096\linewidth]{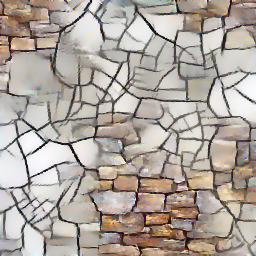}
		\includegraphics[width=0.096\linewidth]{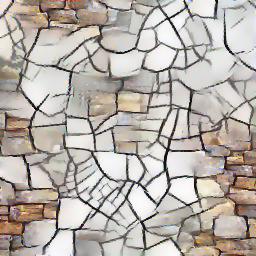}
		\includegraphics[width=0.096\linewidth]{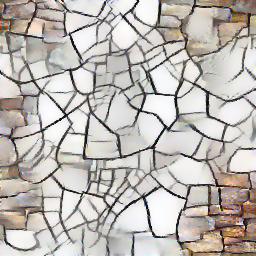}
		\includegraphics[width=0.096\linewidth]{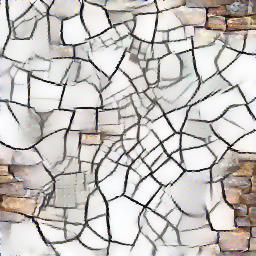}
		\includegraphics[width=0.096\linewidth]{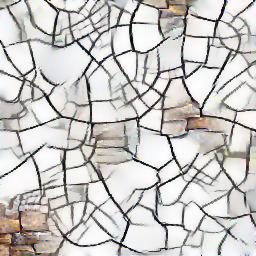}
		\includegraphics[width=0.096\linewidth]{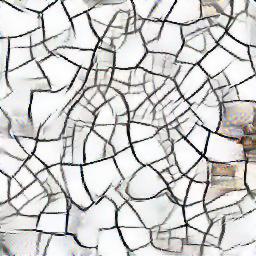}
		\includegraphics[width=0.096\linewidth]{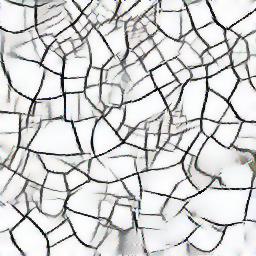}
	}\\	
	\TabHeight{
		\includegraphics[width=0.096\linewidth]{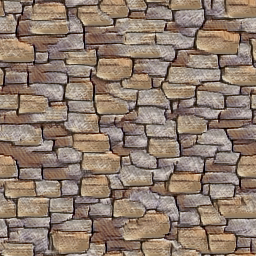}
		\includegraphics[width=0.096\linewidth]{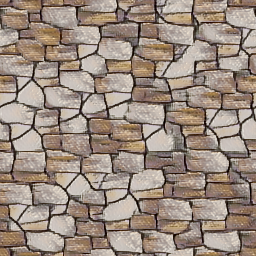}
		\includegraphics[width=0.096\linewidth]{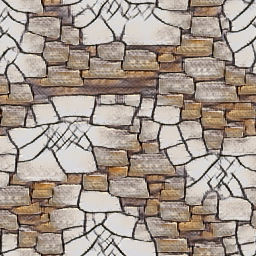}
		\includegraphics[width=0.096\linewidth]{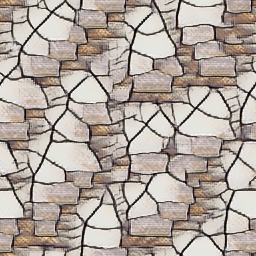}
		\includegraphics[width=0.096\linewidth]{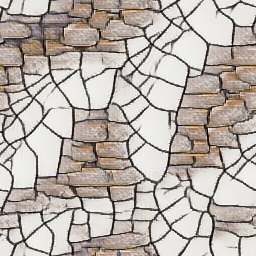}
		\includegraphics[width=0.096\linewidth]{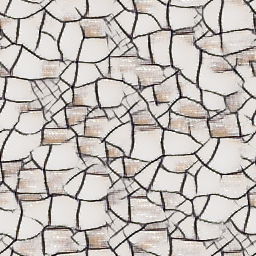}
		\includegraphics[width=0.096\linewidth]{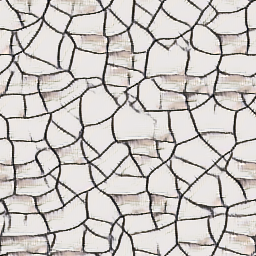}
		\includegraphics[width=0.096\linewidth]{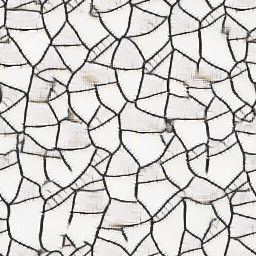}
		\includegraphics[width=0.096\linewidth]{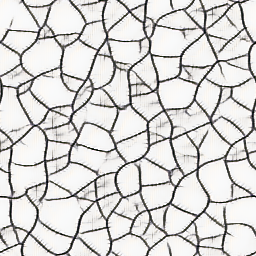}
		\includegraphics[width=0.096\linewidth]{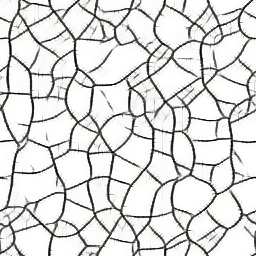}
	}\\
	\TabHeight{
		\includegraphics[width=0.096\linewidth]{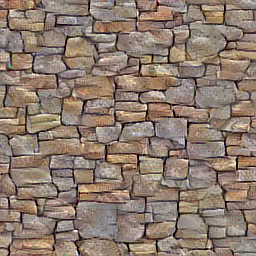}
		\includegraphics[width=0.096\linewidth]{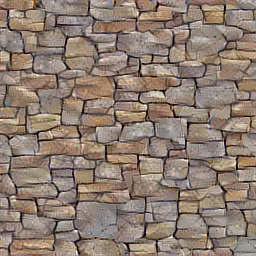}
		\includegraphics[width=0.096\linewidth]{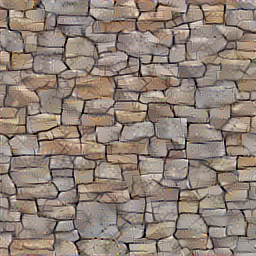}
		\includegraphics[width=0.096\linewidth]{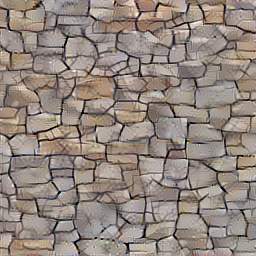}
		\includegraphics[width=0.096\linewidth]{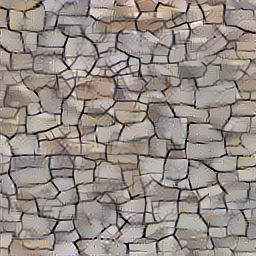}
		\includegraphics[width=0.096\linewidth]{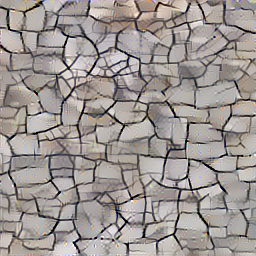}
		\includegraphics[width=0.096\linewidth]{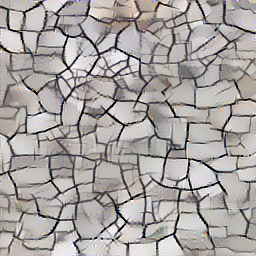}
		\includegraphics[width=0.096\linewidth]{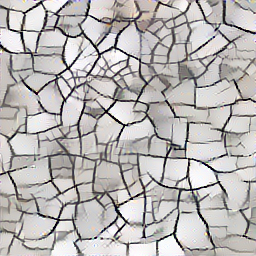}
		\includegraphics[width=0.096\linewidth]{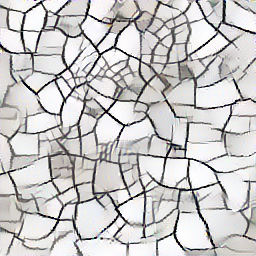}
		\includegraphics[width=0.096\linewidth]{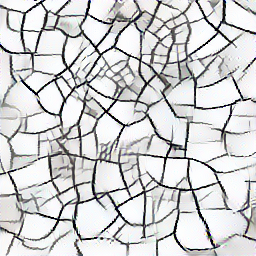}
	}\\
	\TabHeight{
		\includegraphics[width=0.096\linewidth]{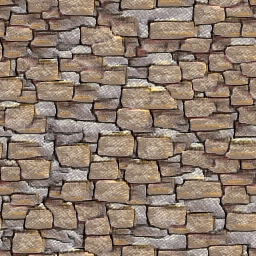}
		\includegraphics[width=0.096\linewidth]{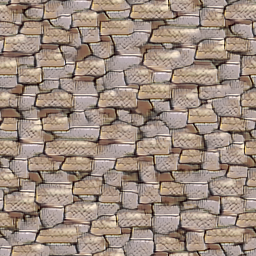}
		\includegraphics[width=0.096\linewidth]{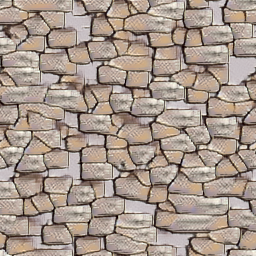}
		\includegraphics[width=0.096\linewidth]{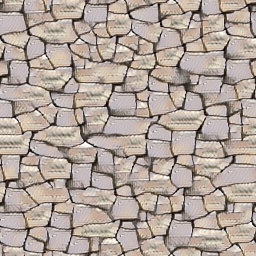}
		\includegraphics[width=0.096\linewidth]{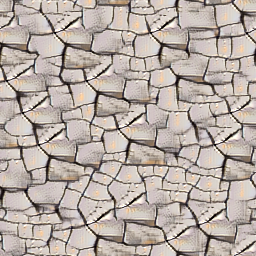}
		\includegraphics[width=0.096\linewidth]{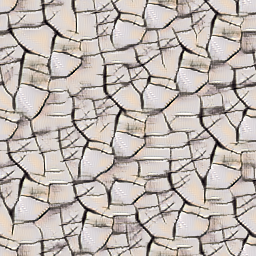}
		\includegraphics[width=0.096\linewidth]{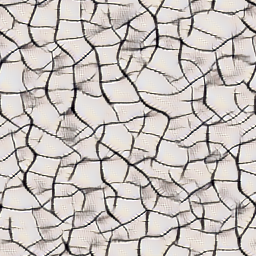}
		\includegraphics[width=0.096\linewidth]{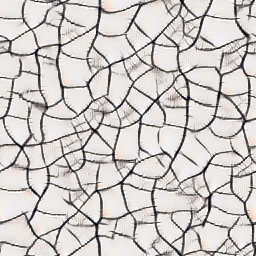}
		\includegraphics[width=0.096\linewidth]{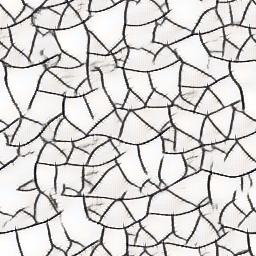}
		\includegraphics[width=0.096\linewidth]{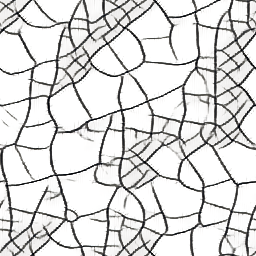}
	}\\	
	\end{center}
\vspace{-3mm}
\caption{Experiments on mixing textures containing complex patterns,
by using Image Melding (1-st row),
DFN (2-rd row),
LIA (3-rd row),
LIA + TextureNet (4-th row),
our scheme + Gram matrix (5-th row),
and our scheme + TextureNet (6-th row).
Note that our scheme, either combining with Gram matrix or with TextureNet, can create more smooth transitions, both in color and texture patterns, between inputs.}
\label{CNN}
\end{figure*}

\begin{figure*}[htb!]
	\begin{center}
	\TabHeight{
		\includegraphics[width=0.096\linewidth]{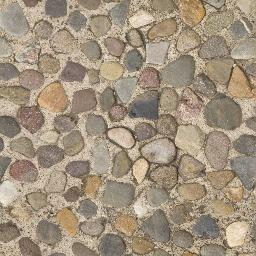}
		\includegraphics[width=0.096\linewidth]{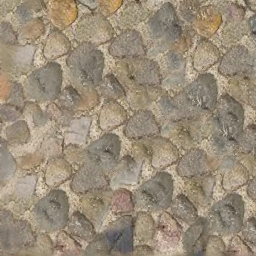}
		\includegraphics[width=0.096\linewidth]{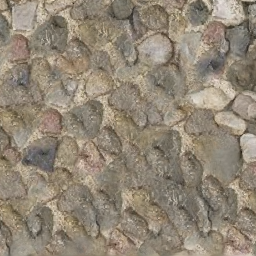}
		\includegraphics[width=0.096\linewidth]{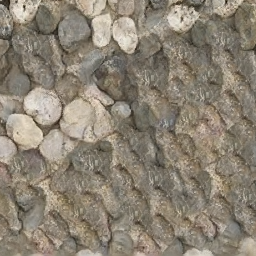}
		\includegraphics[width=0.096\linewidth]{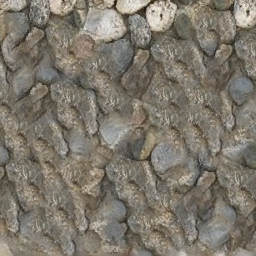}
		\includegraphics[width=0.096\linewidth]{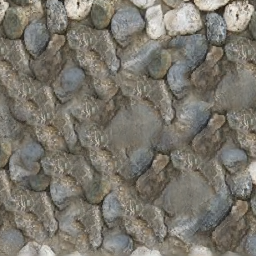}
		\includegraphics[width=0.096\linewidth]{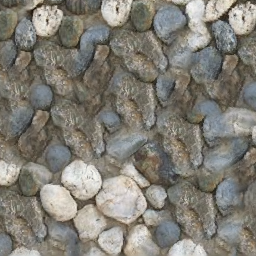}
		\includegraphics[width=0.096\linewidth]{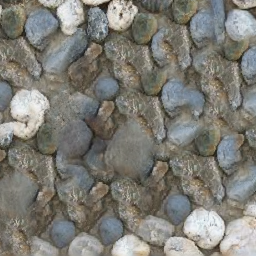}
		\includegraphics[width=0.096\linewidth]{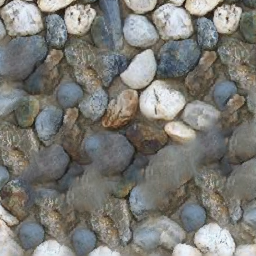}
		\includegraphics[width=0.096\linewidth]{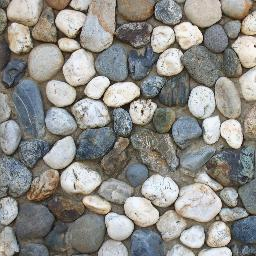}
	}\\
	\TabHeight{
		\includegraphics[width=0.096\linewidth]{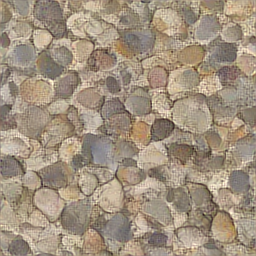}
		\includegraphics[width=0.096\linewidth]{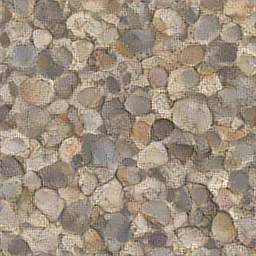}
		\includegraphics[width=0.096\linewidth]{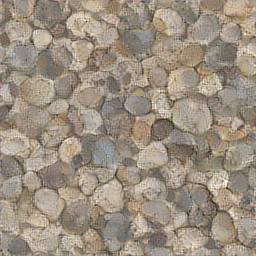}
		\includegraphics[width=0.096\linewidth]{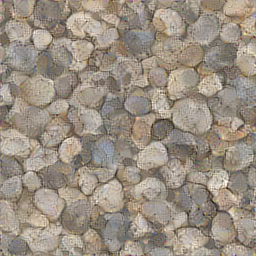}
		\includegraphics[width=0.096\linewidth]{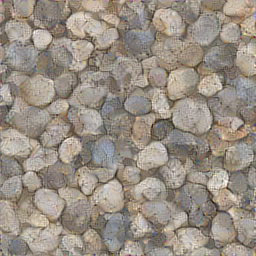}
		\includegraphics[width=0.096\linewidth]{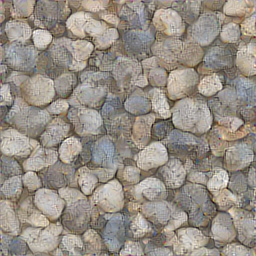}
		\includegraphics[width=0.096\linewidth]{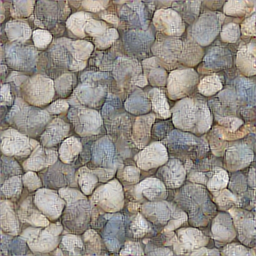}
		\includegraphics[width=0.096\linewidth]{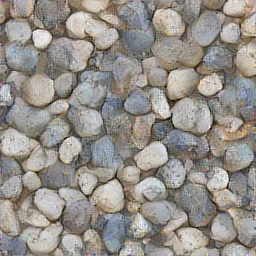}
		\includegraphics[width=0.096\linewidth]{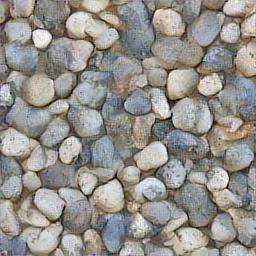}
		\includegraphics[width=0.096\linewidth]{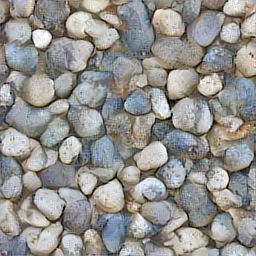}
	}\\

	\TabHeight{
		\includegraphics[width=0.096\linewidth]{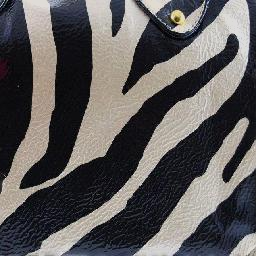}
		\includegraphics[width=0.096\linewidth]{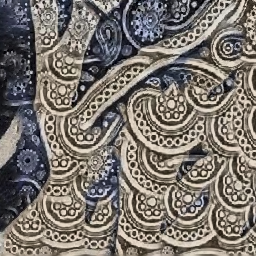}
		\includegraphics[width=0.096\linewidth]{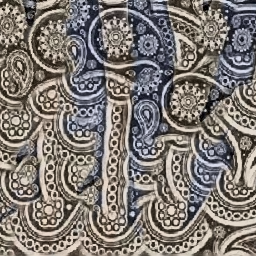}
		\includegraphics[width=0.096\linewidth]{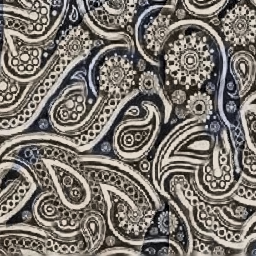}
		\includegraphics[width=0.096\linewidth]{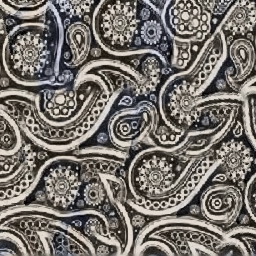}
		\includegraphics[width=0.096\linewidth]{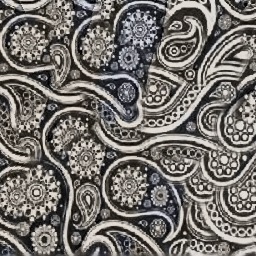}
		\includegraphics[width=0.096\linewidth]{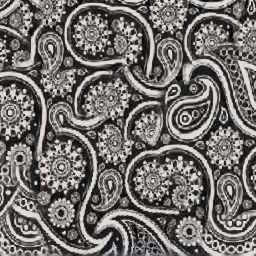}
		\includegraphics[width=0.096\linewidth]{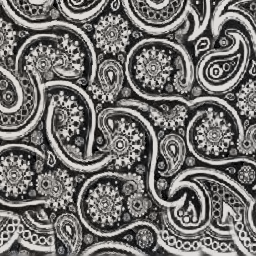}
		\includegraphics[width=0.096\linewidth]{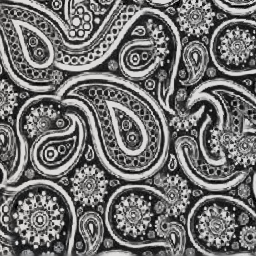}
		\includegraphics[width=0.096\linewidth]{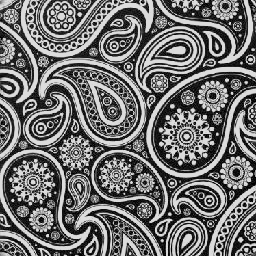}
	}\\
	\TabHeight{
		\includegraphics[width=0.096\linewidth]{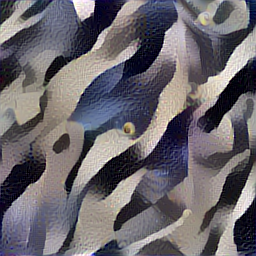}
		\includegraphics[width=0.096\linewidth]{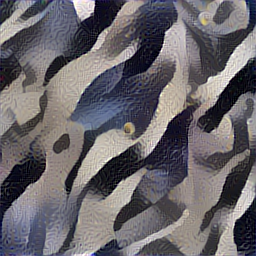}
		\includegraphics[width=0.096\linewidth]{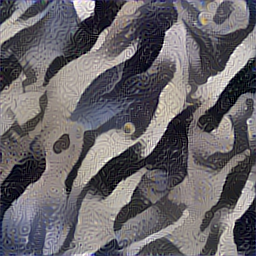}
		\includegraphics[width=0.096\linewidth]{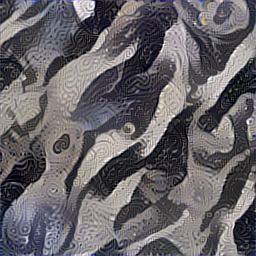}
		\includegraphics[width=0.096\linewidth]{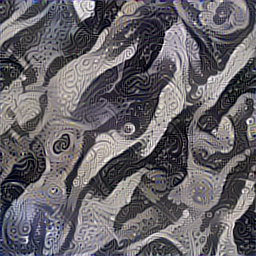}
		\includegraphics[width=0.096\linewidth]{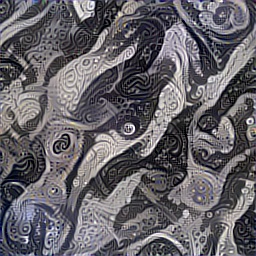}
		\includegraphics[width=0.096\linewidth]{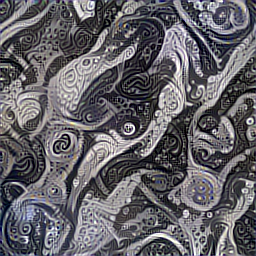}
		\includegraphics[width=0.096\linewidth]{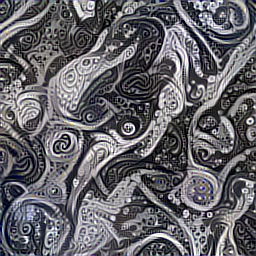}
		\includegraphics[width=0.096\linewidth]{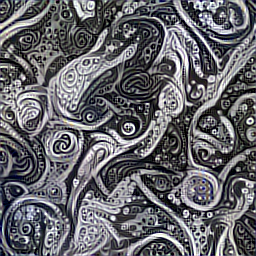}
		\includegraphics[width=0.096\linewidth]{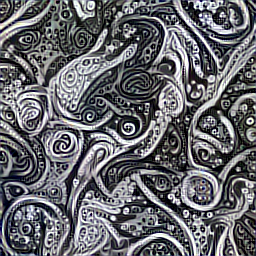}
	}\\
	\end{center}
\vspace{-3mm}
\caption{Comparisons on texture mixing, by using Image Melding (1-st, 3-rd and 5-th row) and using our scheme with Gram matrix (2-nd, 4-th and 6-th row).}
\label{G_only}
\end{figure*}

\begin{figure*}[htb!]
	\begin{center}
	\TabHeight{
		\includegraphics[width=0.096\linewidth]{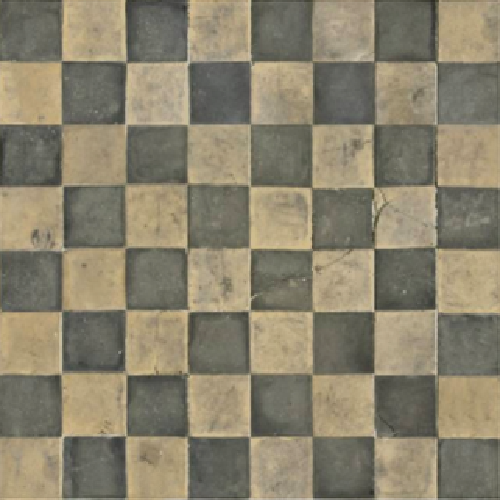}
		\includegraphics[width=0.096\linewidth]{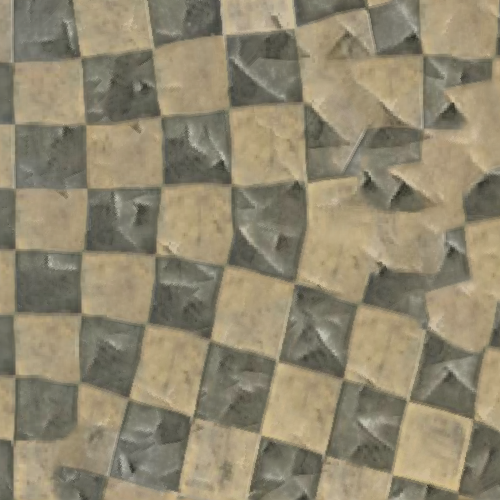}
		\includegraphics[width=0.096\linewidth]{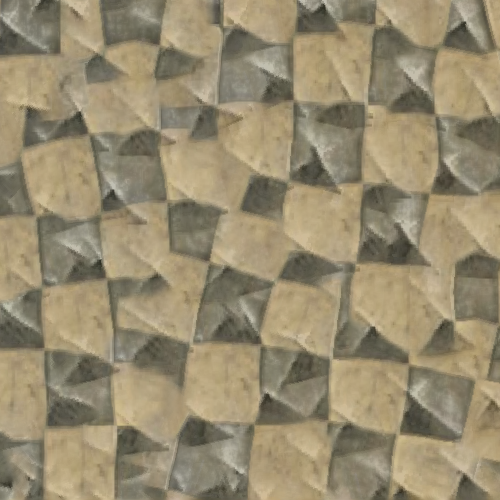}
		\includegraphics[width=0.096\linewidth]{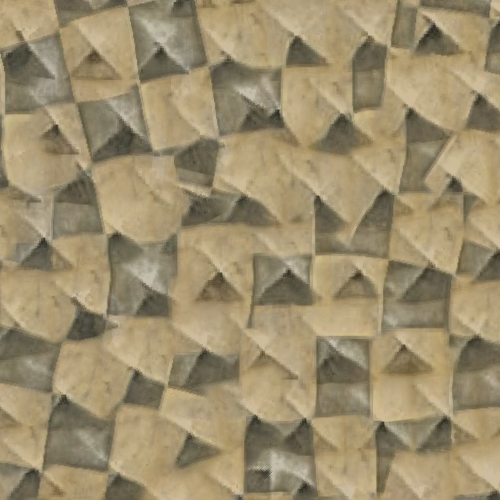}
		\includegraphics[width=0.096\linewidth]{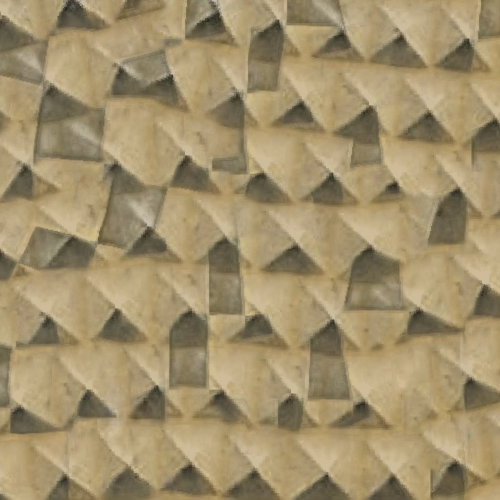}
		\includegraphics[width=0.096\linewidth]{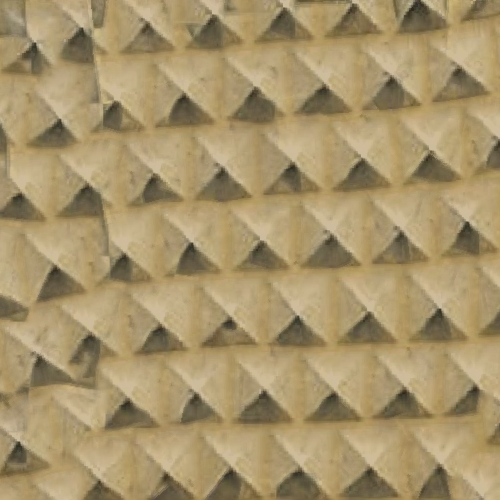}
		\includegraphics[width=0.096\linewidth]{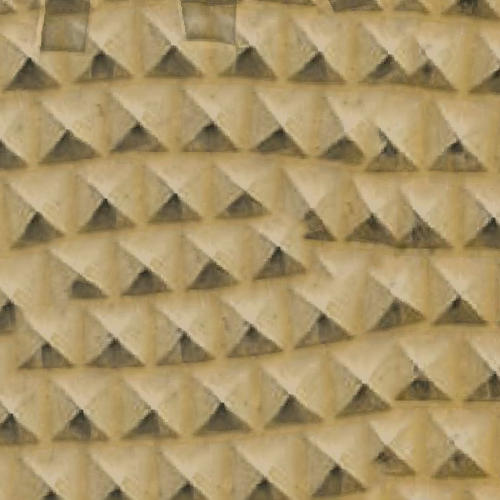}
		\includegraphics[width=0.096\linewidth]{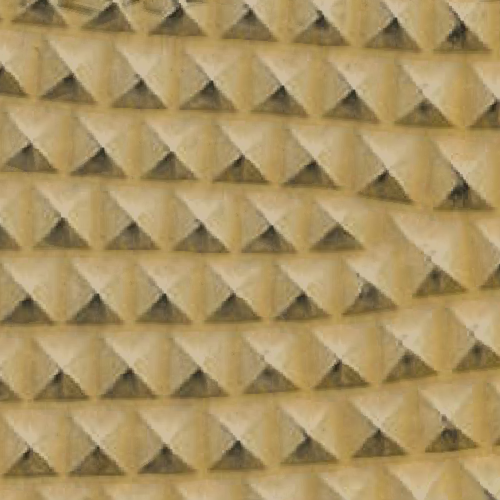}
		\includegraphics[width=0.096\linewidth]{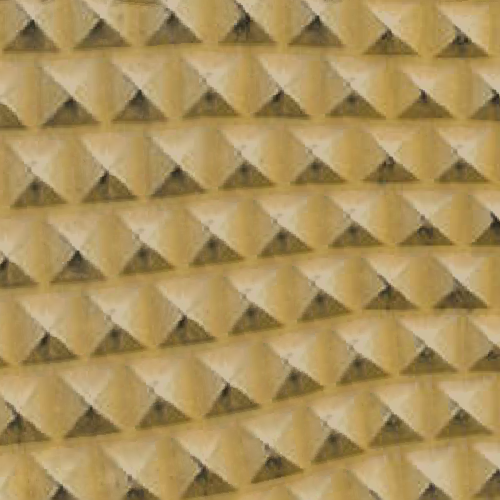}
		\includegraphics[width=0.096\linewidth]{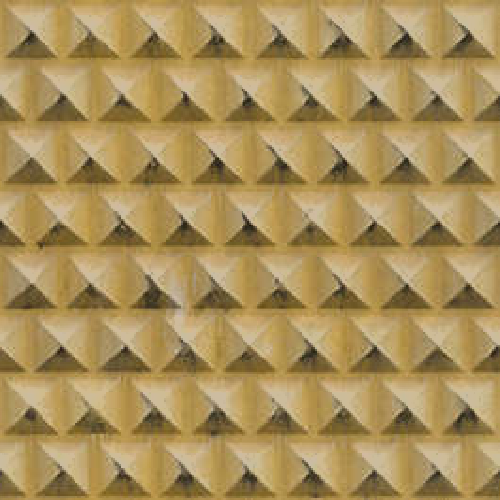}
	}\\
	\TabHeight{
		\includegraphics[width=0.096\linewidth]{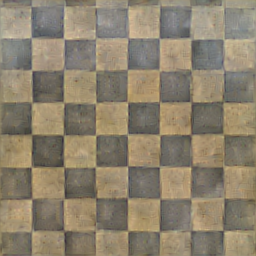}
		\includegraphics[width=0.096\linewidth]{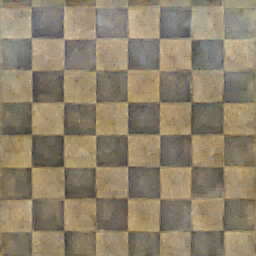}
		\includegraphics[width=0.096\linewidth]{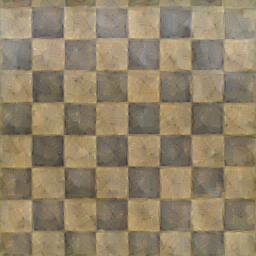}
		\includegraphics[width=0.096\linewidth]{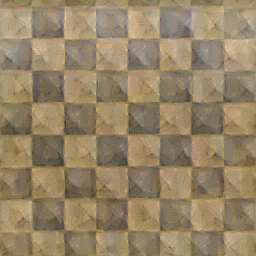}
		\includegraphics[width=0.096\linewidth]{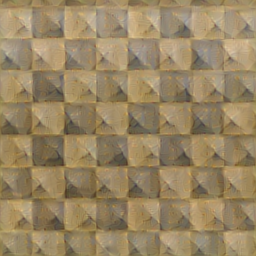}
		\includegraphics[width=0.096\linewidth]{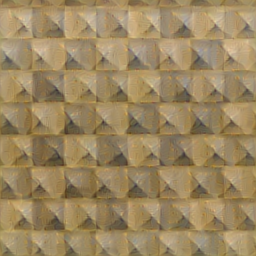}
		\includegraphics[width=0.096\linewidth]{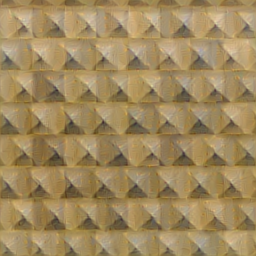}
		\includegraphics[width=0.096\linewidth]{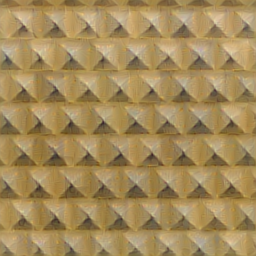}
		\includegraphics[width=0.096\linewidth]{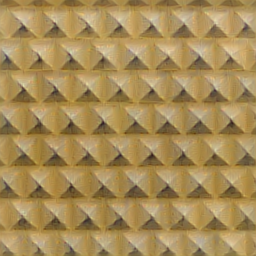}
		\includegraphics[width=0.096\linewidth]{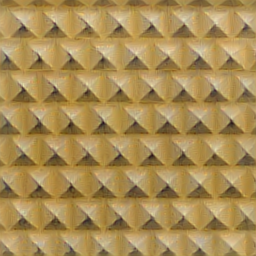}
	}\\
    \TabHeight{
		\includegraphics[width=0.096\linewidth]{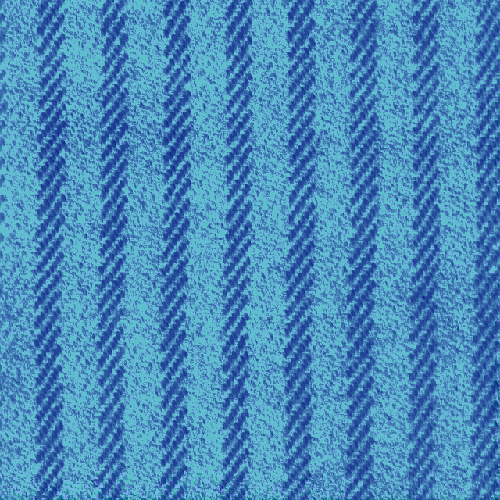}
		\includegraphics[width=0.096\linewidth]{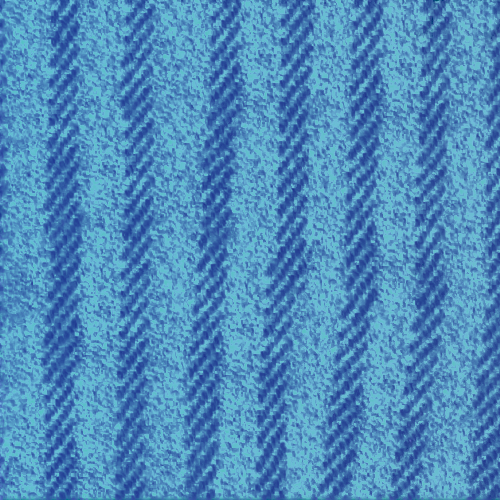}
		\includegraphics[width=0.096\linewidth]{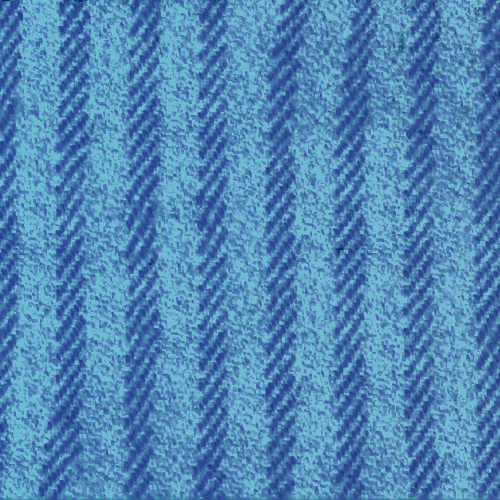}
		\includegraphics[width=0.096\linewidth]{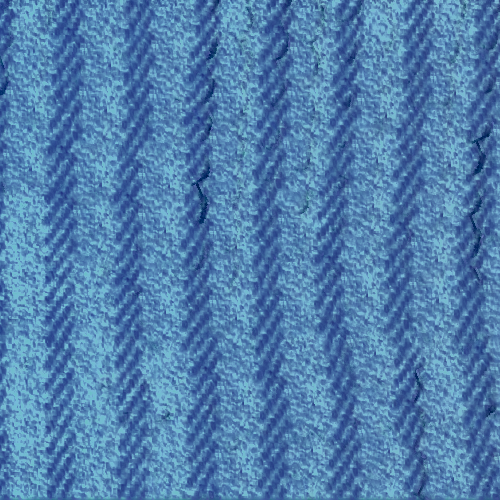}
		\includegraphics[width=0.096\linewidth]{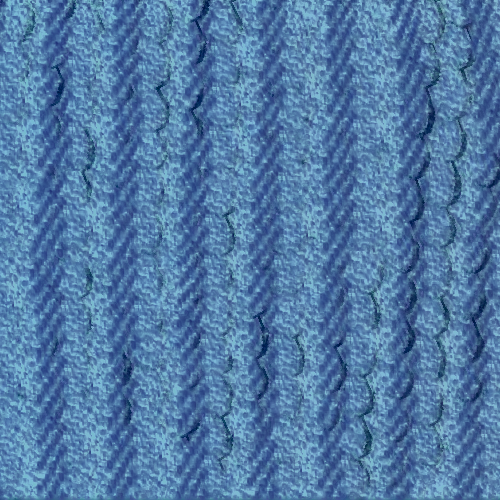}
		\includegraphics[width=0.096\linewidth]{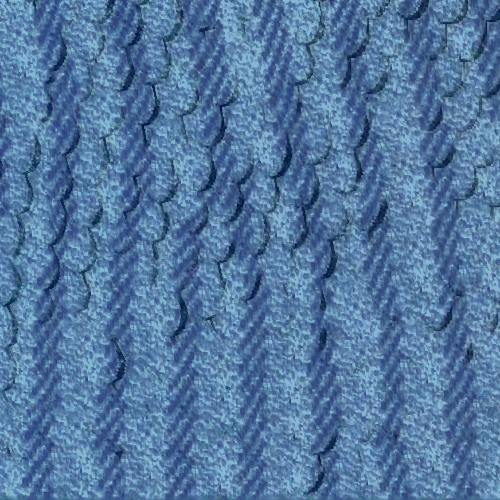}
		\includegraphics[width=0.096\linewidth]{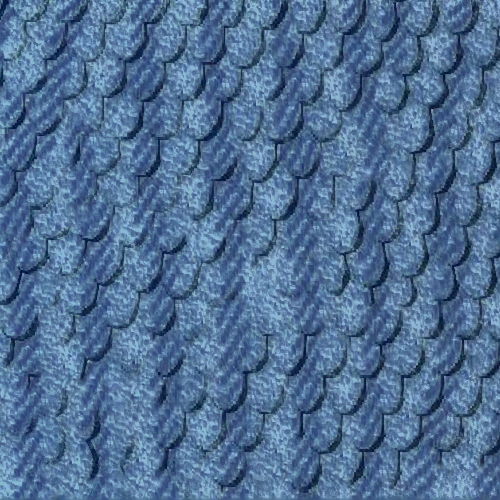}
		\includegraphics[width=0.096\linewidth]{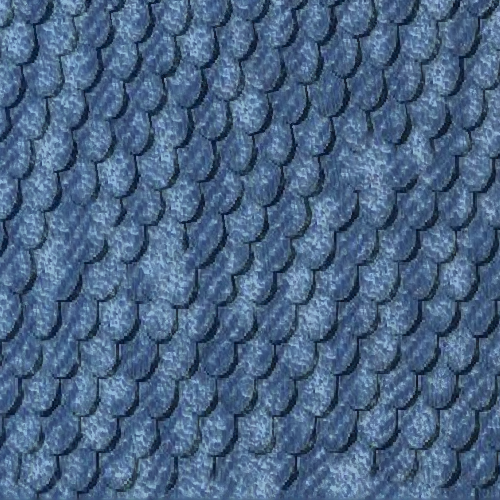}
		\includegraphics[width=0.096\linewidth]{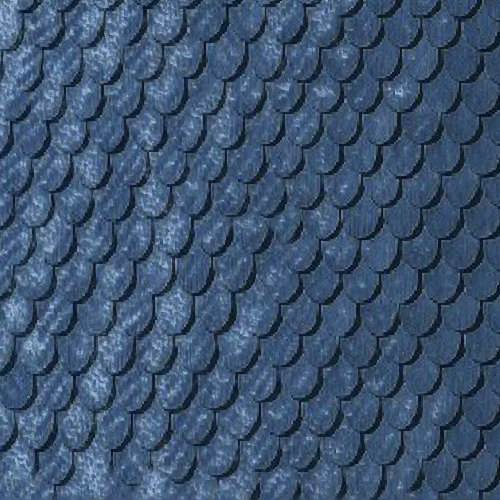}
		\includegraphics[width=0.096\linewidth]{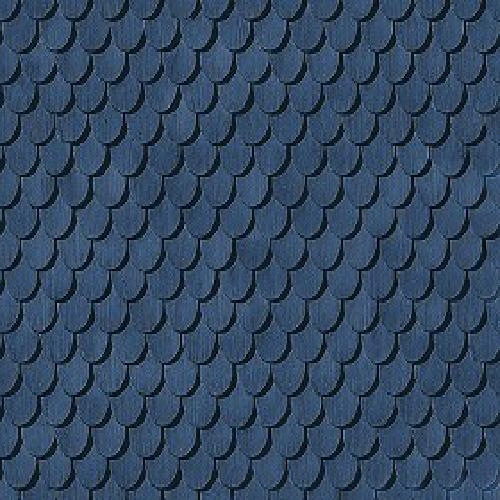}
	}\\
	\TabHeight{
		\includegraphics[width=0.096\linewidth]{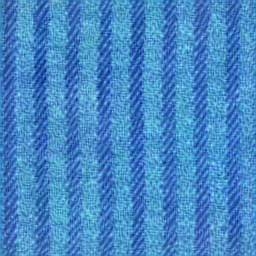}
		\includegraphics[width=0.096\linewidth]{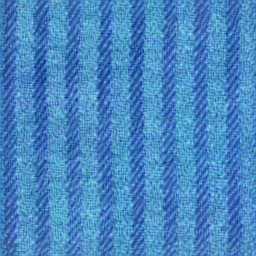}
		\includegraphics[width=0.096\linewidth]{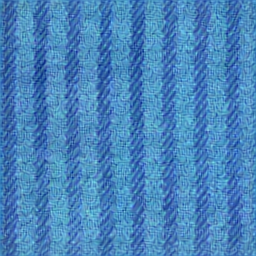}
		\includegraphics[width=0.096\linewidth]{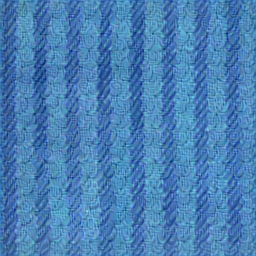}
		\includegraphics[width=0.096\linewidth]{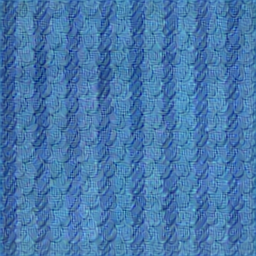}
		\includegraphics[width=0.096\linewidth]{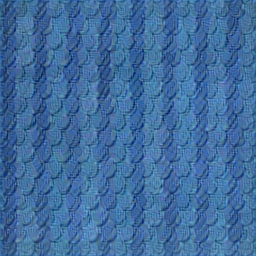}
		\includegraphics[width=0.096\linewidth]{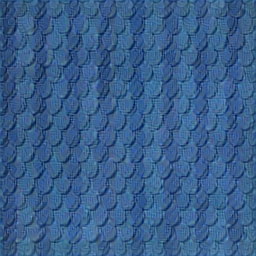}
		\includegraphics[width=0.096\linewidth]{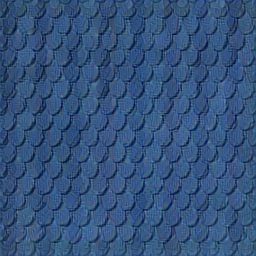}
		\includegraphics[width=0.096\linewidth]{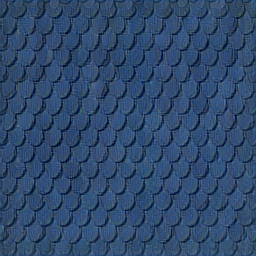}
		\includegraphics[width=0.096\linewidth]{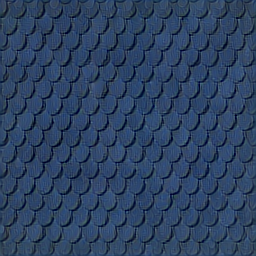}
	}\\
	\end{center}
\caption{Comparisons of mixing periodic textures, by using Image Melding (1-st and 3-rd row) and using our scheme with modified correlation matrix (2-nd and 4-th row). Note that our algorithm can preserve periodicity in every mixed textures.}
\label{G_single}
\end{figure*}

Fig.~\ref{Compare_All} displays the results of mixing two micro textures which can mostly be described by Gaussian model. In this experiment, we compare our algorithm with the GaussTexton~\cite{xia2014synthesizing}, which is specifically designed for micro-texture mixing, and Image Melding~\cite{darabi2012image}. Note that the shape of the grass is completely missed by GaussTexton, while our method can represent it well. Image melding can indeed generate comparable textures, but it produces new structures, {\em i.e.} vertical strips, that are in neither of the input exemplars. Linear mixing algorithm leads to low-quality results, with different textures jointed together patchwisely. The combination of TextureNet and our method gives comparable results, and the output textures are generated instantly. Linear interpolation algorithm (LIA) always leads to poor quality results, with or without TextureNet, where different textures are joint together in patch-wise.

Fig.~\ref{CNN} shows the results on mixing two textures containing more geometric structures, such as sharpe edges which go far beyond Gaussian models. In this experiment, we compared our method with Image Melding, which used patch match for morphing structural textures, and diversified feed-forward networks (DFN)~\cite{li2017diversified}. For DFN, we directly used their trained 60-texture model in this experiment. Because their method is based on Gram matrices, for fair comparisons, we compared our Gram-based methods with theirs. Observe that the mixing procedure of DFN indeed generated new textures, however, some of these textures are not visually similar to neither of the input exemplars. In contrast, our method can produce considerably better results, creating smooth transitions both in color and texture patterns from one to the other input. Image melding failed to generate intermediate textures in this experiments.

Fig.~\ref{G_only} presents more comparisons between our Gram-based algorithm and Image Melding~\cite{darabi2012image}. For the pebbles textures, our Gram-based mixing algorithm can mix the edges and the shapes of pebbles simultaneously, and create smooth transitions from one texture to the other. Image Melding can also create such transition, but it generate obviously repeated patterns, {\em i.e.} some pebbles in mixed textures are completely the same. For the crack textures, our algorithm creates mixed textures where two cracks gradually merged into each other, but in the results of Image Melding, the structures in the first textures are mostly ignored. Our algorithm also performs better on the third pair of textures, where delicate structures are homogeneously mixed together, while Image Melding fails to represent the strip patterns.

Fig.~\ref{G_single} presents more comparison between our correlation based algorithm and Image Melding on periodic textures. For these textures, Image Melding creates intermediate textures which are no longer periodic. On the contrary, our correlation-based algorithm is able to preserve periodicity in every mixed textures.

As a conclusion, our method by interpolating deep statistics via Gaussian models provides a flexible scheme to mix textures with different texture synthesis models using deep neural networks, such as Gatys's method~\cite{gatys2015texture} and its variants~\cite{liu2016texture,sendik2017deep}, and TextureNet~\cite{ulyanov2017improved}. It can produce smooth transitions both in color and texture patterns.

\subsection{Style Morphing}
\begin{figure*}[htb!]
	\begin{center}
		\includegraphics[width=0.75\linewidth]{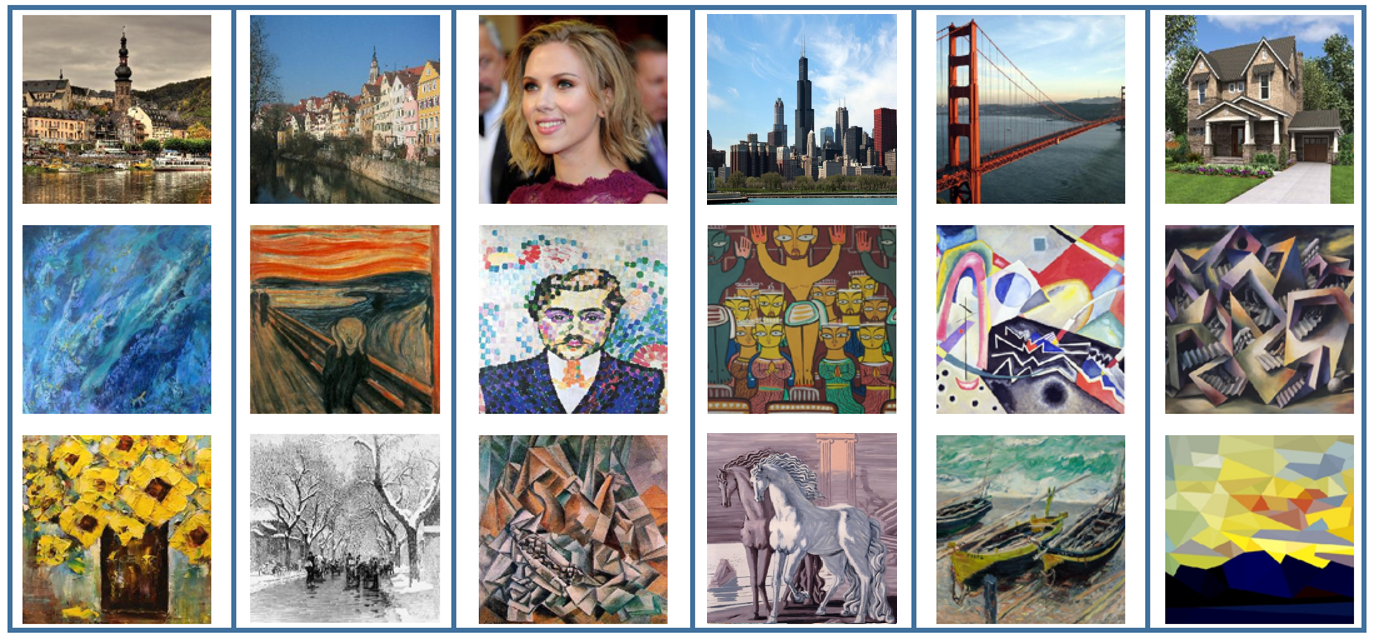}
	\end{center}
\vspace{-3mm}
\caption{Photos and pairs of style images (in column) used for style morphing.}
\label{Styles}
\end{figure*}

In this experiment, we extend our texture mixing to style morphing, our goal is to create ``intermediate'' styles between different styles, in another word, to create smooth transitions between stylish photos. We use Jonson's feed-forward structure~\cite{johnson2016perceptual} together with instance normalization~\cite{ulyanov2017improved}. We set style layers as \texttt{relu1\_1}, \texttt{relu2\_1}, \texttt{relu3\_1} and \texttt{relu4\_1}, content layer as \texttt{relu4\_2}. Style weight is set to 5. All other parameters are left as default. The photo and style images are shown in Fig.~\ref{Styles}.  We compare our result with Dumoulin's algorithm \cite{dumoulin2016learned}.

\begin{figure*}[htb!]
	\begin{center}
	\TabHeight{
		\includegraphics[width=0.096\linewidth]{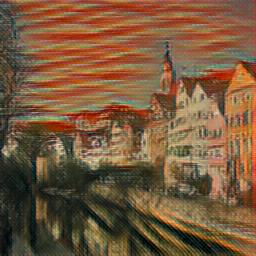}
		\includegraphics[width=0.096\linewidth]{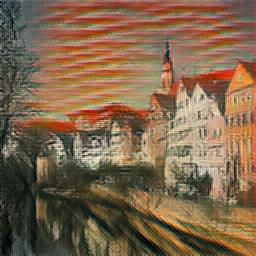}
		\includegraphics[width=0.096\linewidth]{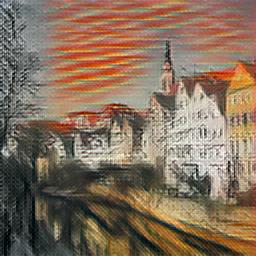}
		\includegraphics[width=0.096\linewidth]{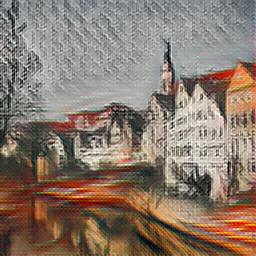}
		\includegraphics[width=0.096\linewidth]{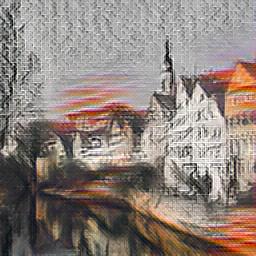}
		\includegraphics[width=0.096\linewidth]{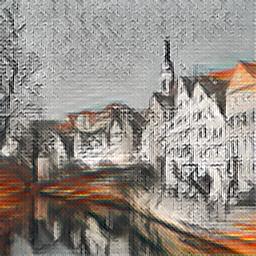}
		\includegraphics[width=0.096\linewidth]{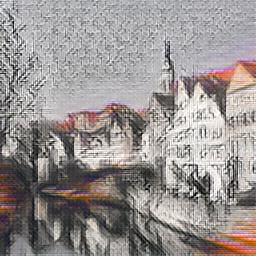}
		\includegraphics[width=0.096\linewidth]{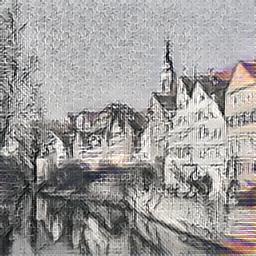}
		\includegraphics[width=0.096\linewidth]{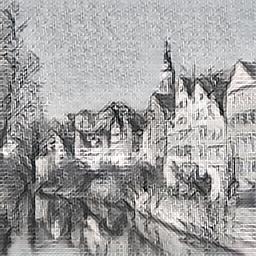}
		\includegraphics[width=0.096\linewidth]{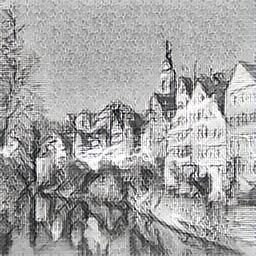}
	}\\
	\TabHeight{
		\includegraphics[width=0.096\linewidth]{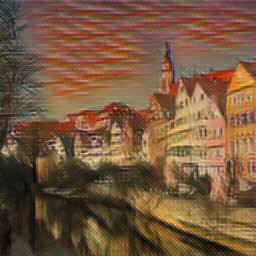}
		\includegraphics[width=0.096\linewidth]{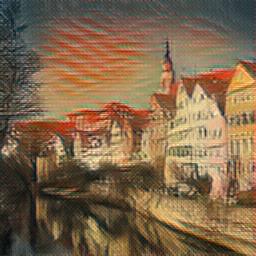}
		\includegraphics[width=0.096\linewidth]{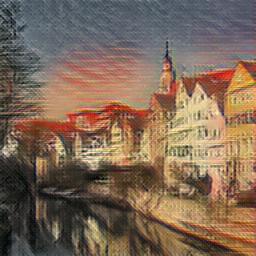}
		\includegraphics[width=0.096\linewidth]{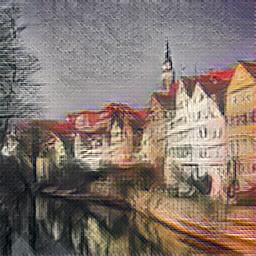}
		\includegraphics[width=0.096\linewidth]{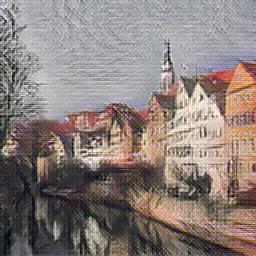}
		\includegraphics[width=0.096\linewidth]{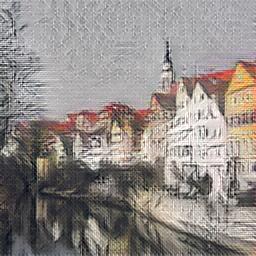}
		\includegraphics[width=0.096\linewidth]{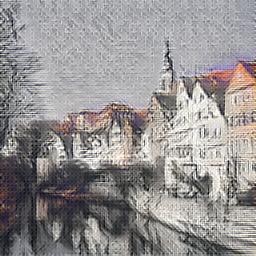}
		\includegraphics[width=0.096\linewidth]{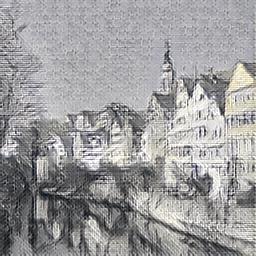}
		\includegraphics[width=0.096\linewidth]{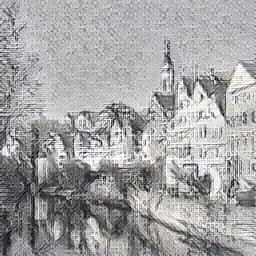}
		\includegraphics[width=0.096\linewidth]{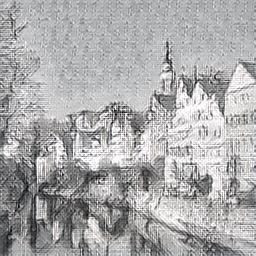}
	}\\
	\TabHeight{
		\includegraphics[width=0.096\linewidth]{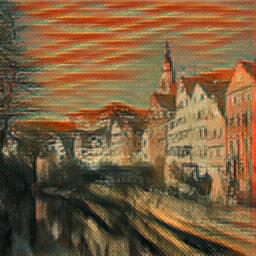}
		\includegraphics[width=0.096\linewidth]{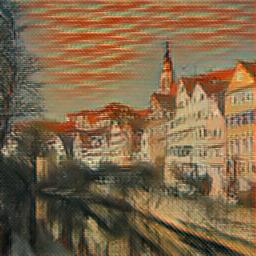}
		\includegraphics[width=0.096\linewidth]{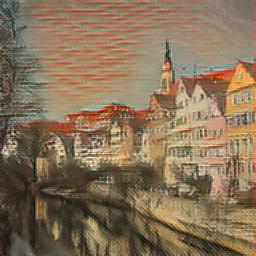}
		\includegraphics[width=0.096\linewidth]{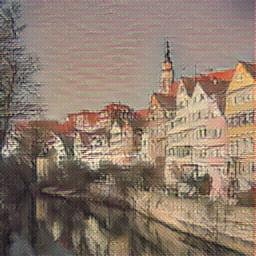}
		\includegraphics[width=0.096\linewidth]{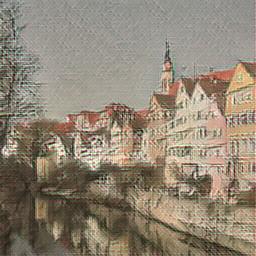}
		\includegraphics[width=0.096\linewidth]{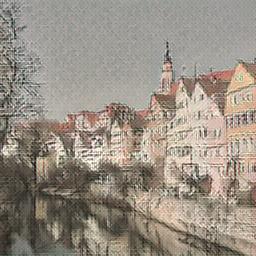}
		\includegraphics[width=0.096\linewidth]{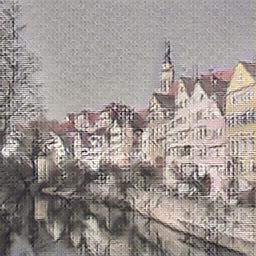}
		\includegraphics[width=0.096\linewidth]{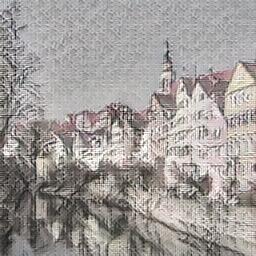}
		\includegraphics[width=0.096\linewidth]{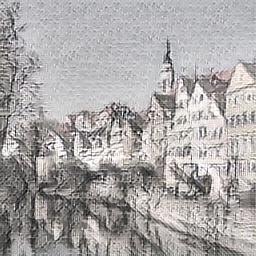}
		\includegraphics[width=0.096\linewidth]{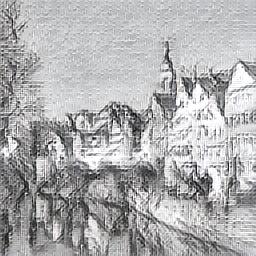}
	}\\

	\TabHeight{
		\includegraphics[width=0.096\linewidth]{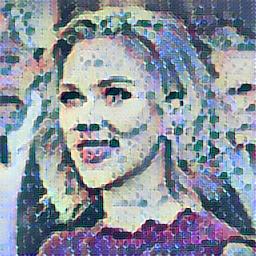}
		\includegraphics[width=0.096\linewidth]{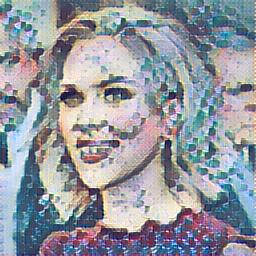}
		\includegraphics[width=0.096\linewidth]{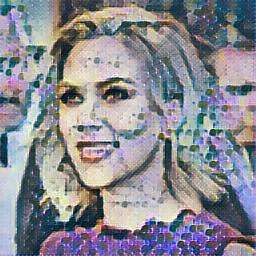}
		\includegraphics[width=0.096\linewidth]{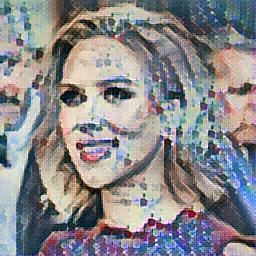}
		\includegraphics[width=0.096\linewidth]{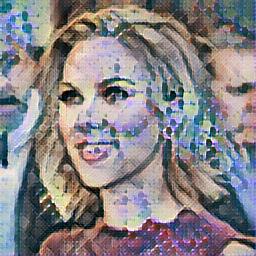}
		\includegraphics[width=0.096\linewidth]{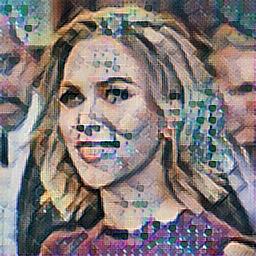}
		\includegraphics[width=0.096\linewidth]{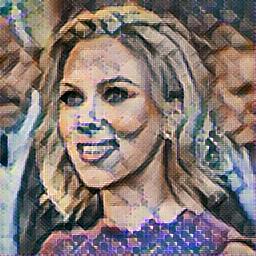}
		\includegraphics[width=0.096\linewidth]{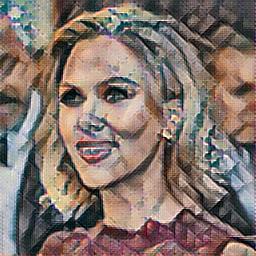}
		\includegraphics[width=0.096\linewidth]{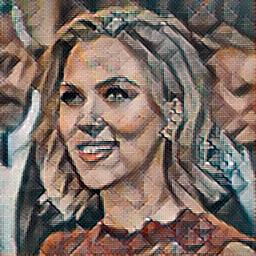}
		\includegraphics[width=0.096\linewidth]{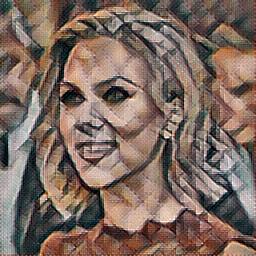}
	}\\
	\TabHeight{
		\includegraphics[width=0.096\linewidth]{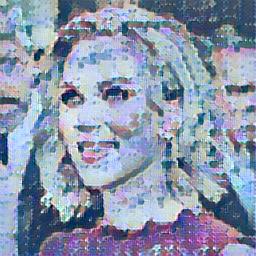}
		\includegraphics[width=0.096\linewidth]{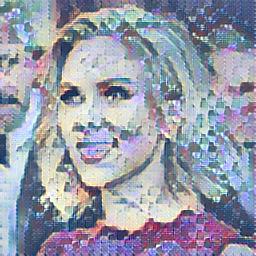}
		\includegraphics[width=0.096\linewidth]{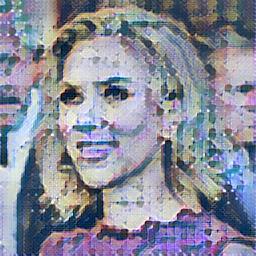}
		\includegraphics[width=0.096\linewidth]{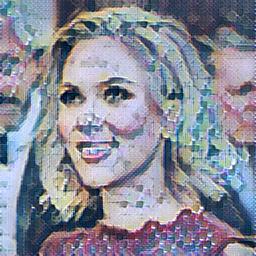}
		\includegraphics[width=0.096\linewidth]{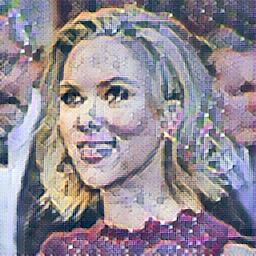}
		\includegraphics[width=0.096\linewidth]{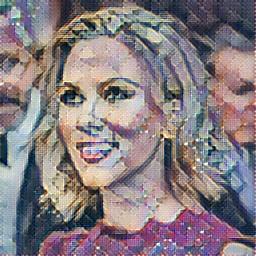}
		\includegraphics[width=0.096\linewidth]{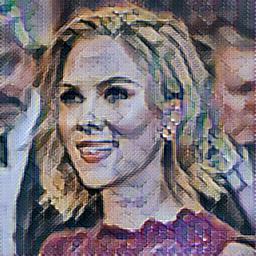}
		\includegraphics[width=0.096\linewidth]{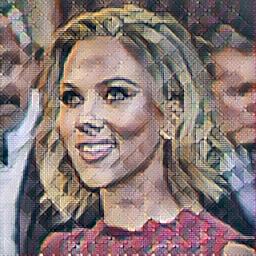}
		\includegraphics[width=0.096\linewidth]{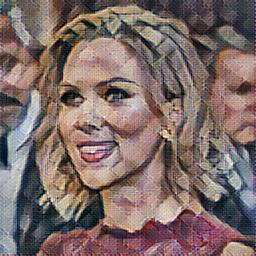}
		\includegraphics[width=0.096\linewidth]{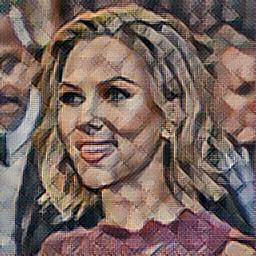}
	}\\
	\TabHeight{
		\includegraphics[width=0.096\linewidth]{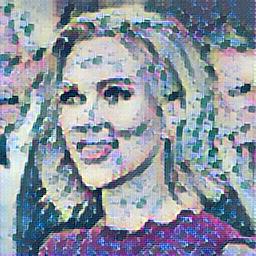}
		\includegraphics[width=0.096\linewidth]{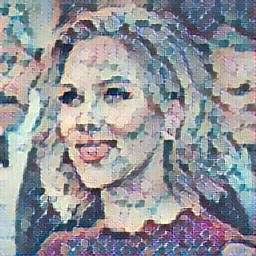}
		\includegraphics[width=0.096\linewidth]{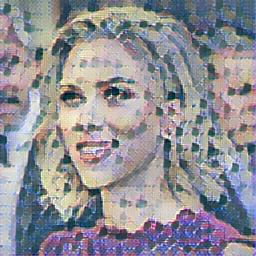}
		\includegraphics[width=0.096\linewidth]{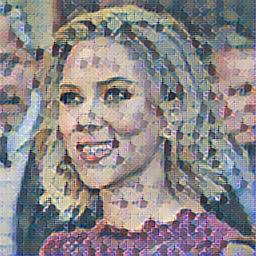}
		\includegraphics[width=0.096\linewidth]{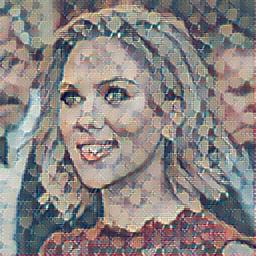}
		\includegraphics[width=0.096\linewidth]{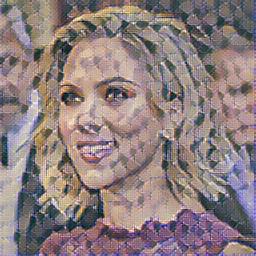}
		\includegraphics[width=0.096\linewidth]{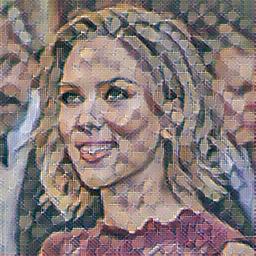}
		\includegraphics[width=0.096\linewidth]{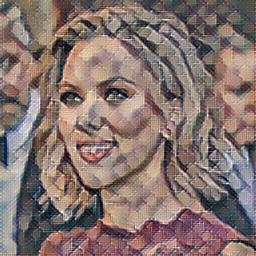}
		\includegraphics[width=0.096\linewidth]{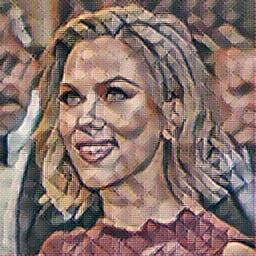}
		\includegraphics[width=0.096\linewidth]{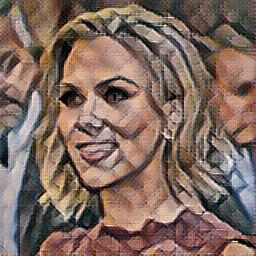}
	}\\
	\end{center}
\vspace{-3mm}
\caption{Comparisons of style morphing results, by using linear algorithm (1-st and 4-th row), using our scheme without lag constraint (2-nd and 5th row), and using our scheme with lag constraint (3-rd and 6-rd row). Note that our results with lag constraint technique (see text for details) gives better result to mix styles homogeneously.}
\label{LAG}
\end{figure*}

In our experiments, we propose to use a technique called \emph{lag constraint} to morph styles. Specifically,
we mix feature maps at \texttt{pool1}, \texttt{pool2} and \texttt{pool3} layer and propagate these mixed feature maps to \texttt{relu2\_1}, \texttt{relu3\_1} and \texttt{relu4\_1} layers respectively, and use these feature maps at ``relu'' layers as constrain.
Results generated with or without lag constraint are showed in Fig.~\ref{LAG}, the results with lag constraint have higher visual quality, as it can mixed different styles more smoothly without artifacts.

\begin{figure*}[htb!]
	\begin{center}	
	\TabHeight{
		\includegraphics[width=0.096\linewidth]{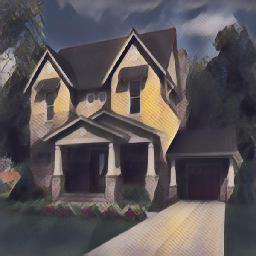}
		\includegraphics[width=0.096\linewidth]{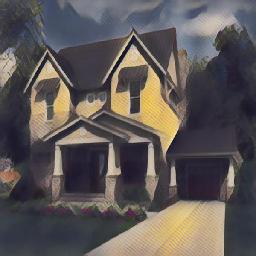}
		\includegraphics[width=0.096\linewidth]{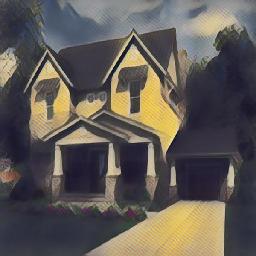}
		\includegraphics[width=0.096\linewidth]{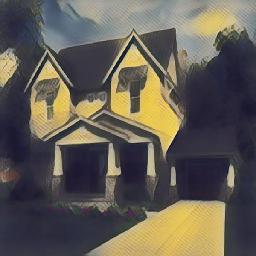}
		\includegraphics[width=0.096\linewidth]{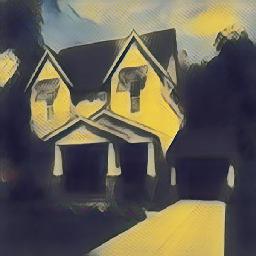}
		\includegraphics[width=0.096\linewidth]{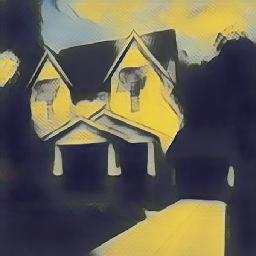}
		\includegraphics[width=0.096\linewidth]{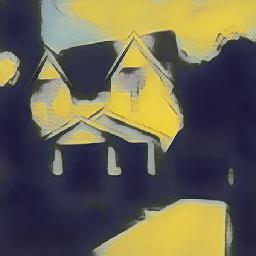}
		\includegraphics[width=0.096\linewidth]{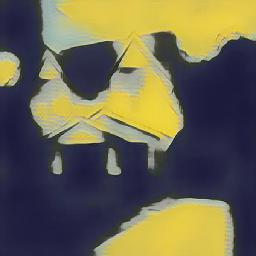}
		\includegraphics[width=0.096\linewidth]{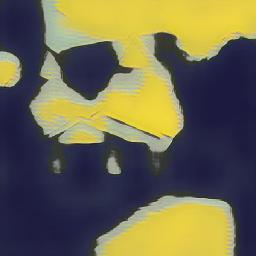}
		\includegraphics[width=0.096\linewidth]{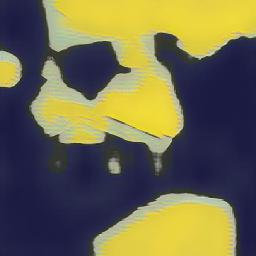}
	}\\
	\TabHeight{
		\includegraphics[width=0.096\linewidth]{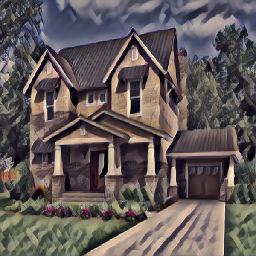}
		\includegraphics[width=0.096\linewidth]{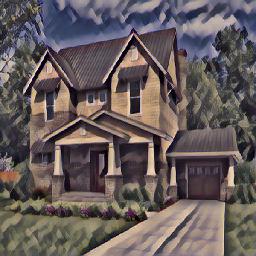}
		\includegraphics[width=0.096\linewidth]{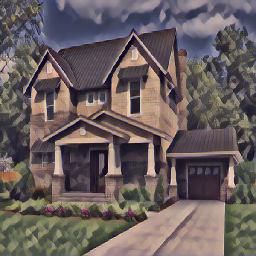}
		\includegraphics[width=0.096\linewidth]{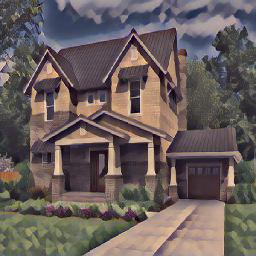}
		\includegraphics[width=0.096\linewidth]{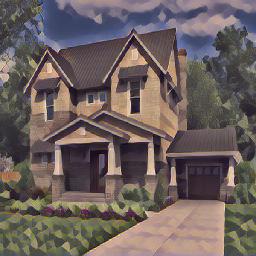}
		\includegraphics[width=0.096\linewidth]{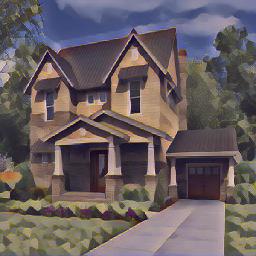}
		\includegraphics[width=0.096\linewidth]{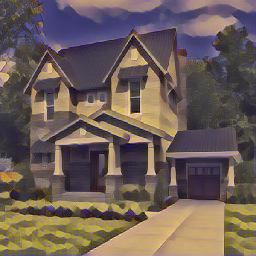}
		\includegraphics[width=0.096\linewidth]{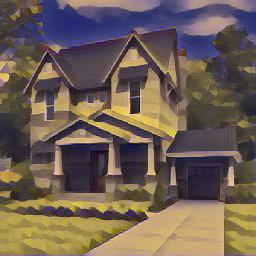}
		\includegraphics[width=0.096\linewidth]{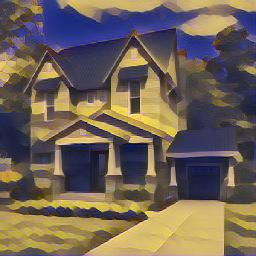}
		\includegraphics[width=0.096\linewidth]{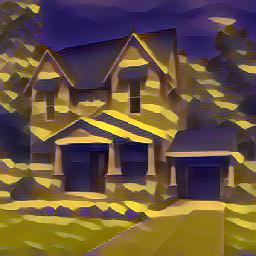}
	}\\
	\TabHeight{
		\includegraphics[width=0.096\linewidth]{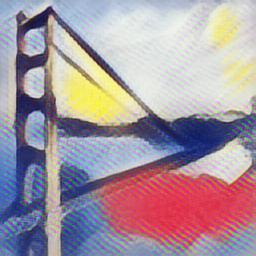}
		\includegraphics[width=0.096\linewidth]{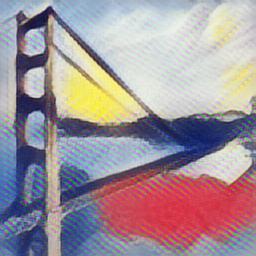}
		\includegraphics[width=0.096\linewidth]{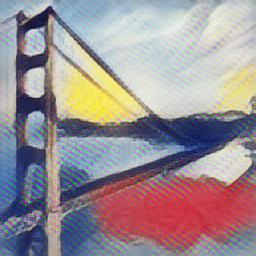}
		\includegraphics[width=0.096\linewidth]{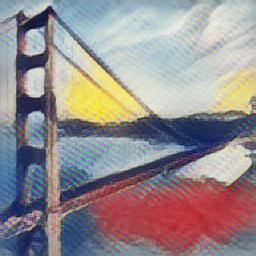}
		\includegraphics[width=0.096\linewidth]{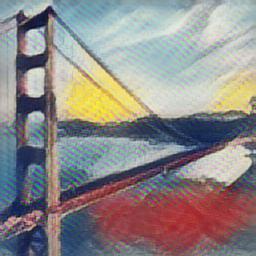}
		\includegraphics[width=0.096\linewidth]{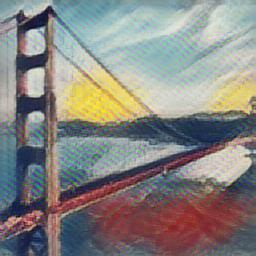}
		\includegraphics[width=0.096\linewidth]{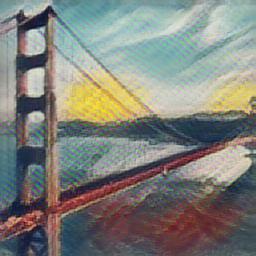}
		\includegraphics[width=0.096\linewidth]{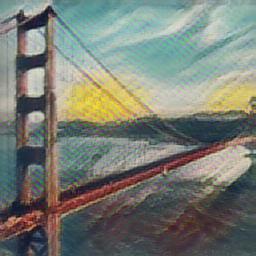}
		\includegraphics[width=0.096\linewidth]{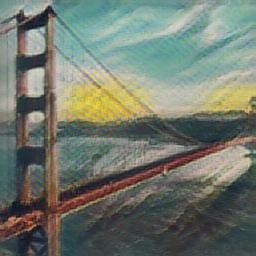}
		\includegraphics[width=0.096\linewidth]{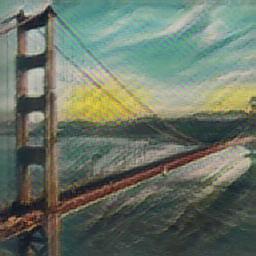}
	}\\
	\TabHeight{
		\includegraphics[width=0.096\linewidth]{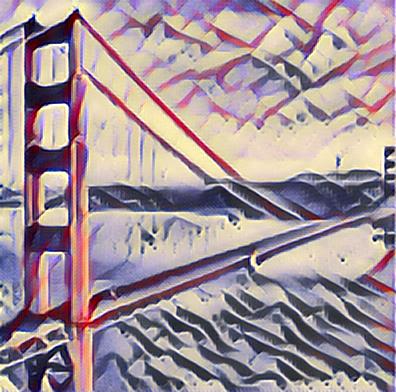}
		\includegraphics[width=0.096\linewidth]{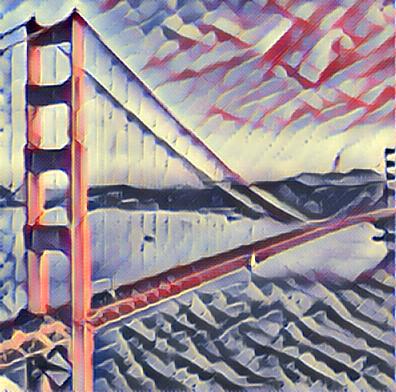}
		\includegraphics[width=0.096\linewidth]{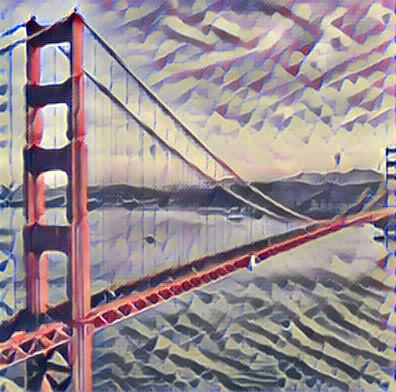}
		\includegraphics[width=0.096\linewidth]{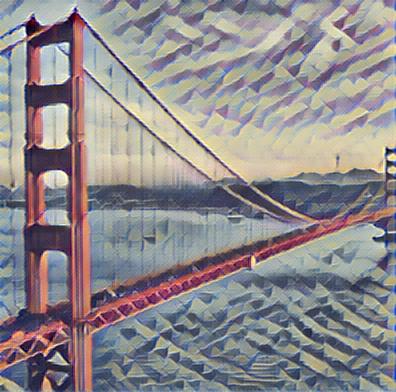}
		\includegraphics[width=0.096\linewidth]{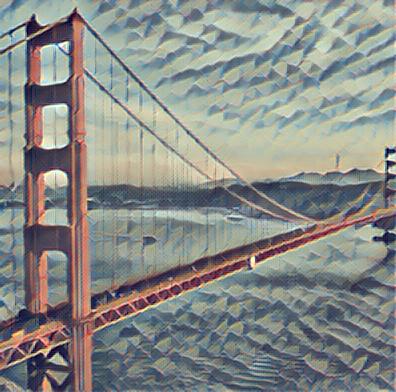}
		\includegraphics[width=0.096\linewidth]{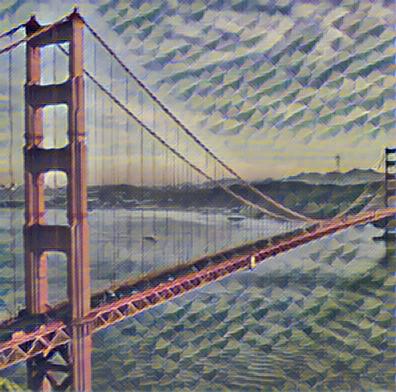}
		\includegraphics[width=0.096\linewidth]{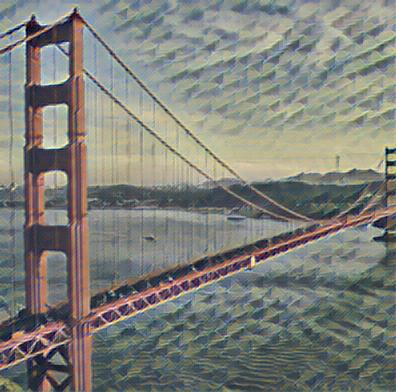}
		\includegraphics[width=0.096\linewidth]{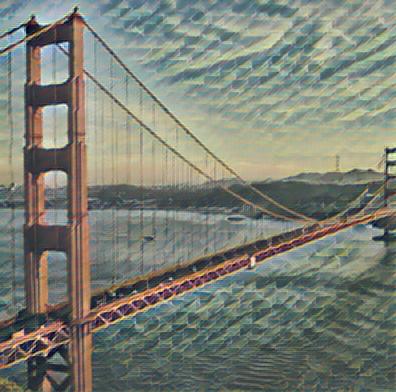}
		\includegraphics[width=0.096\linewidth]{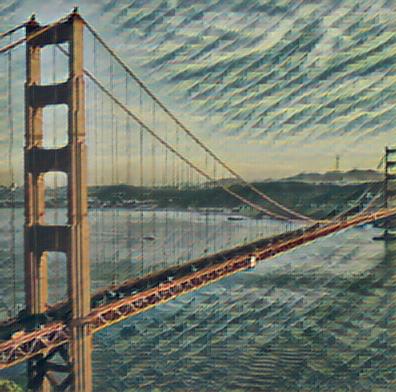}
		\includegraphics[width=0.096\linewidth]{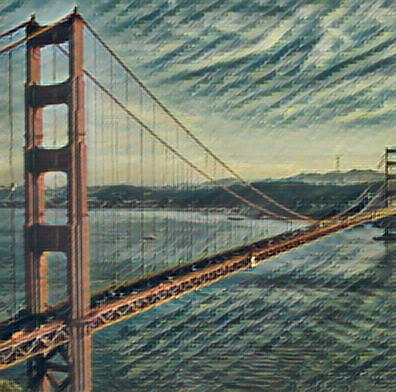}
	}\\
	\TabHeight{
		\includegraphics[width=0.096\linewidth]{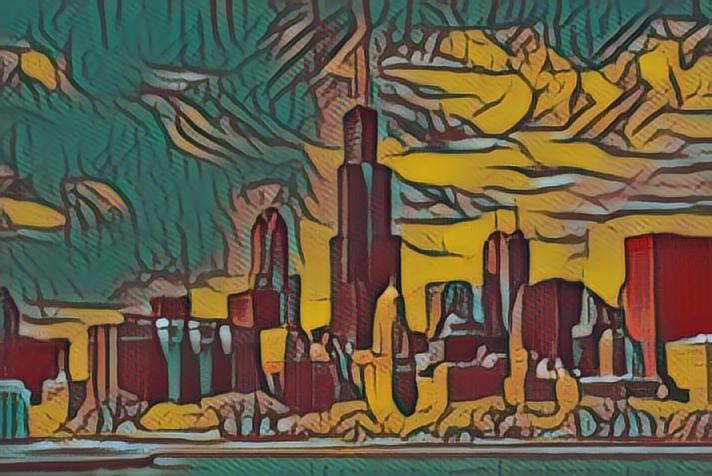}
		\includegraphics[width=0.096\linewidth]{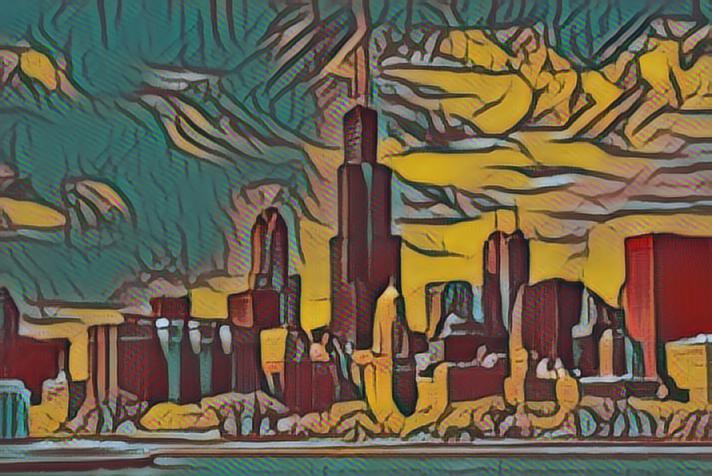}
		\includegraphics[width=0.096\linewidth]{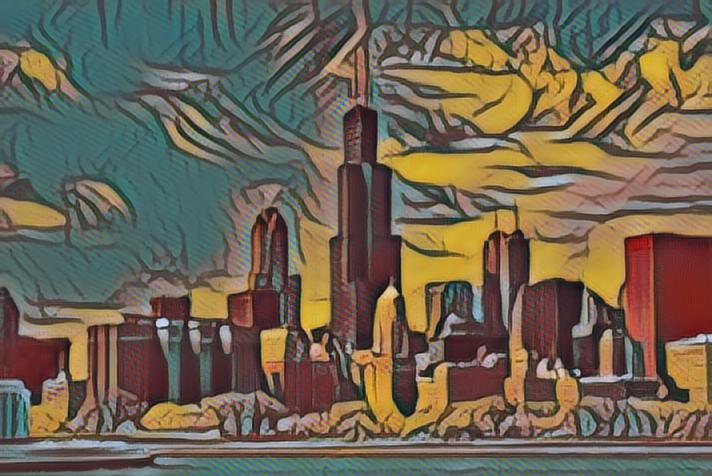}
		\includegraphics[width=0.096\linewidth]{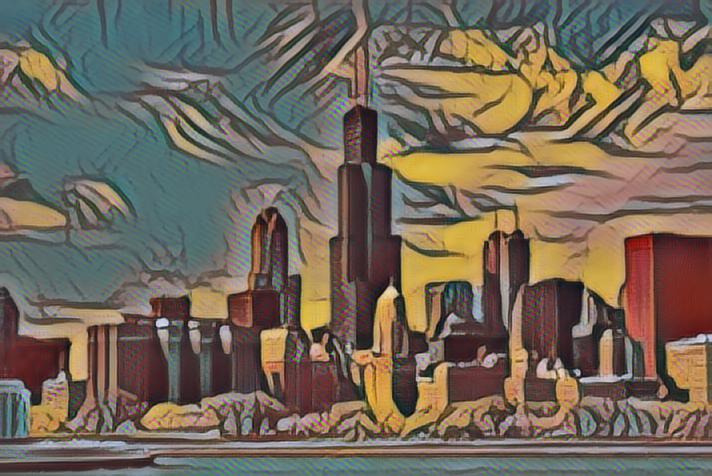}
		\includegraphics[width=0.096\linewidth]{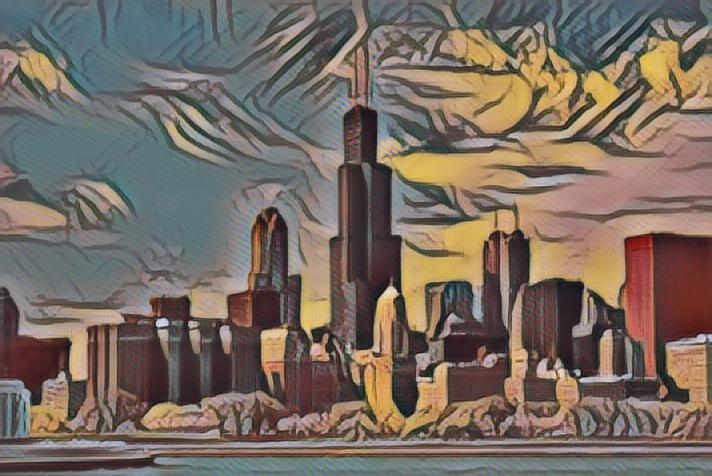}
		\includegraphics[width=0.096\linewidth]{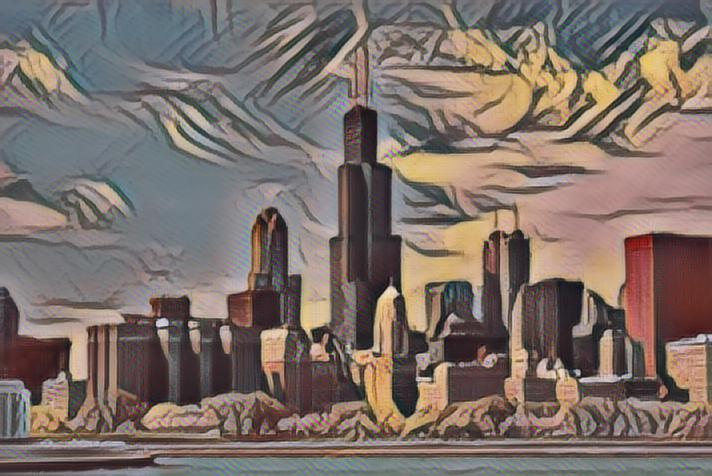}
		\includegraphics[width=0.096\linewidth]{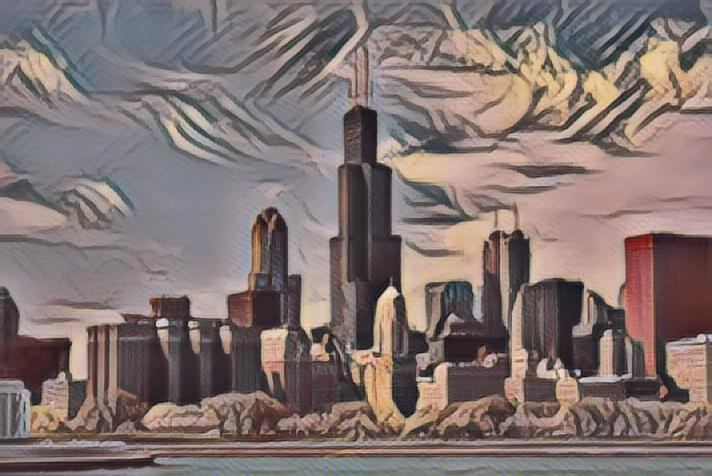}
		\includegraphics[width=0.096\linewidth]{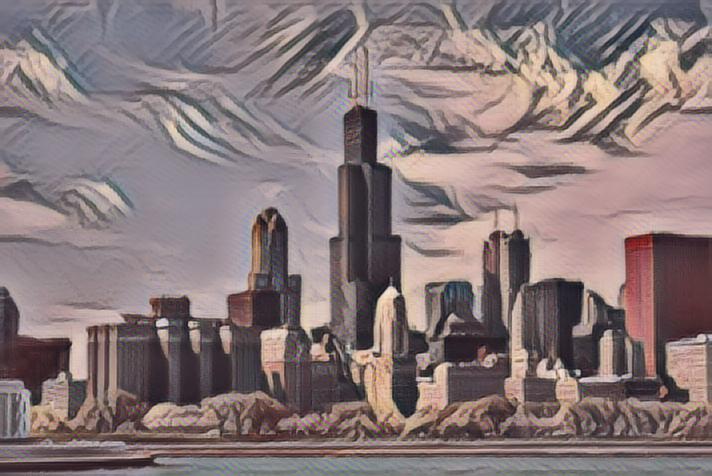}
		\includegraphics[width=0.096\linewidth]{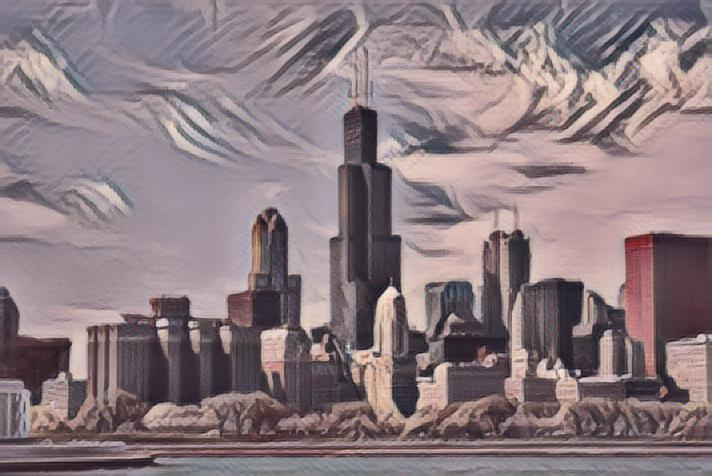}
		\includegraphics[width=0.096\linewidth]{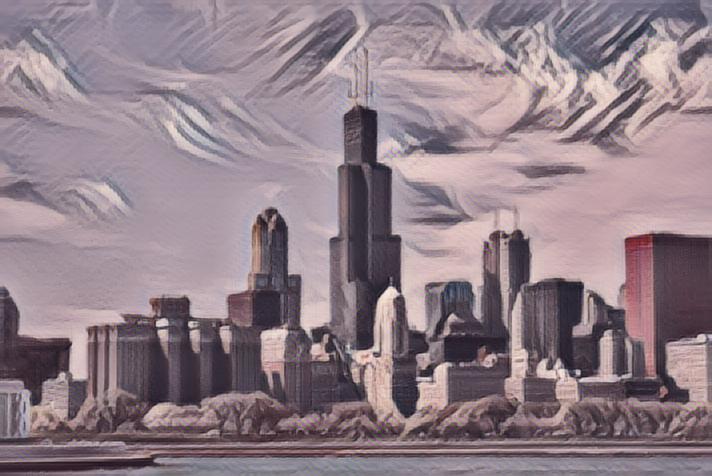}
	}\\
	\TabHeight{
		\includegraphics[width=0.096\linewidth]{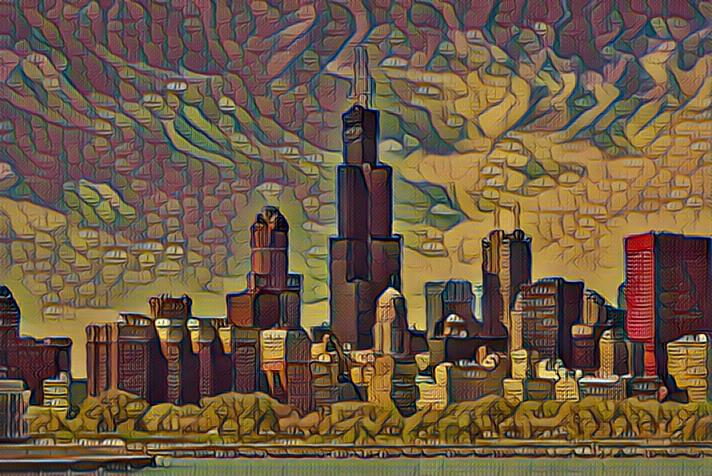}
		\includegraphics[width=0.096\linewidth]{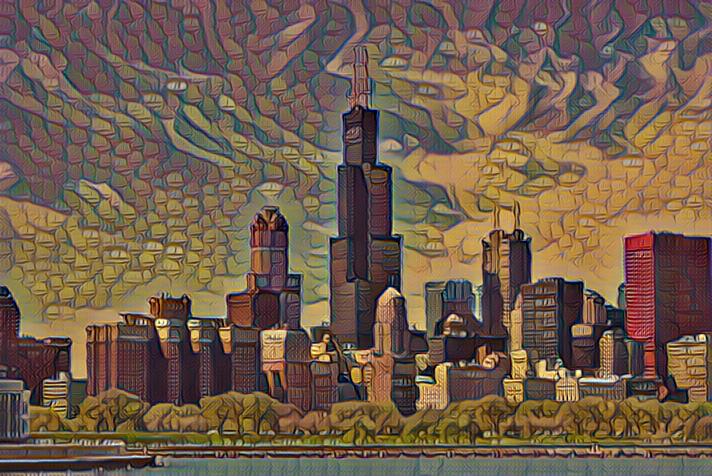}
		\includegraphics[width=0.096\linewidth]{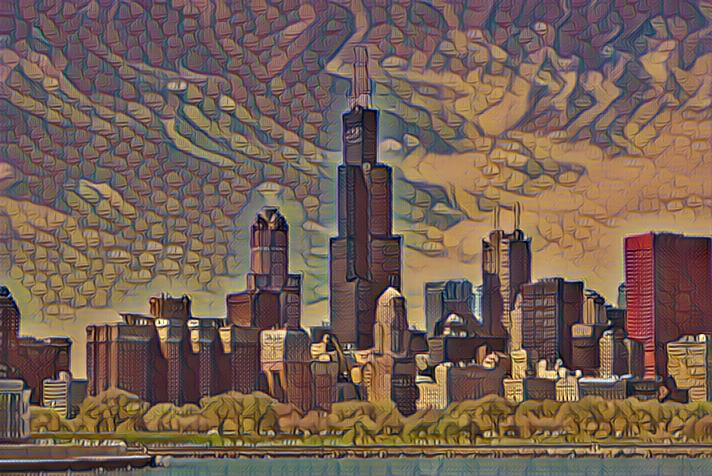}
		\includegraphics[width=0.096\linewidth]{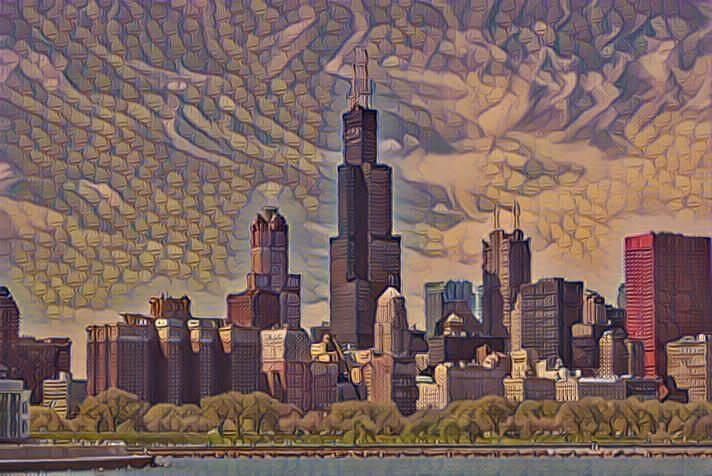}
		\includegraphics[width=0.096\linewidth]{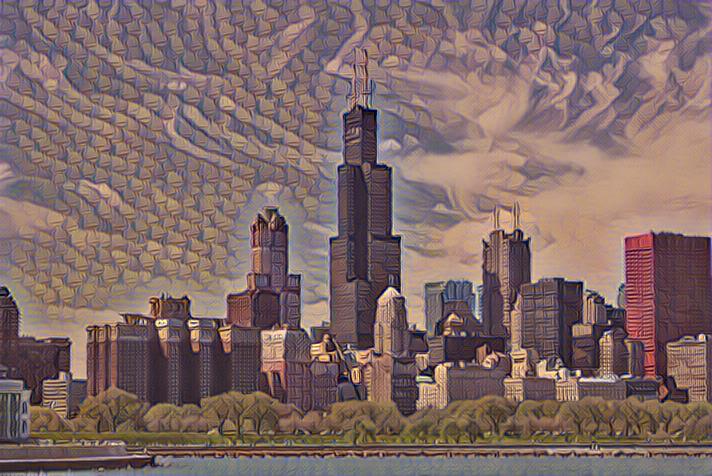}
		\includegraphics[width=0.096\linewidth]{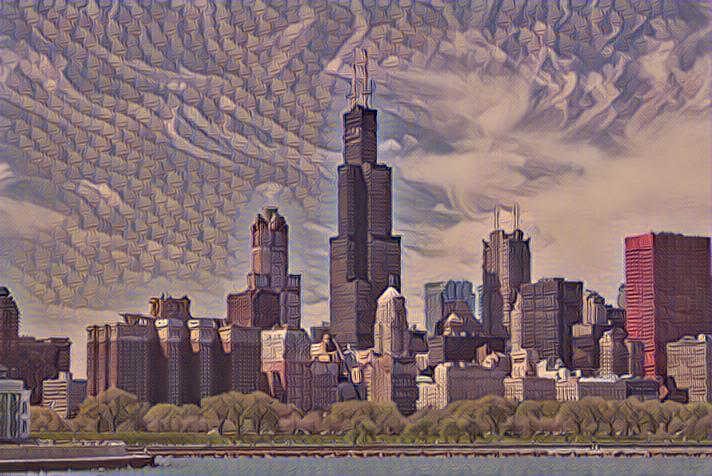}
		\includegraphics[width=0.096\linewidth]{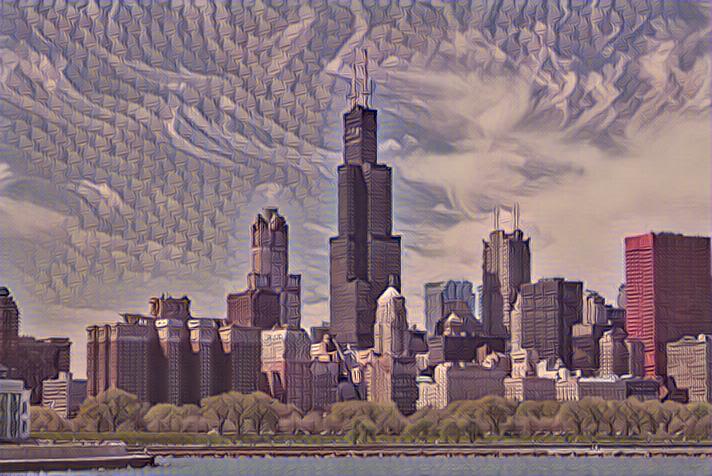}
		\includegraphics[width=0.096\linewidth]{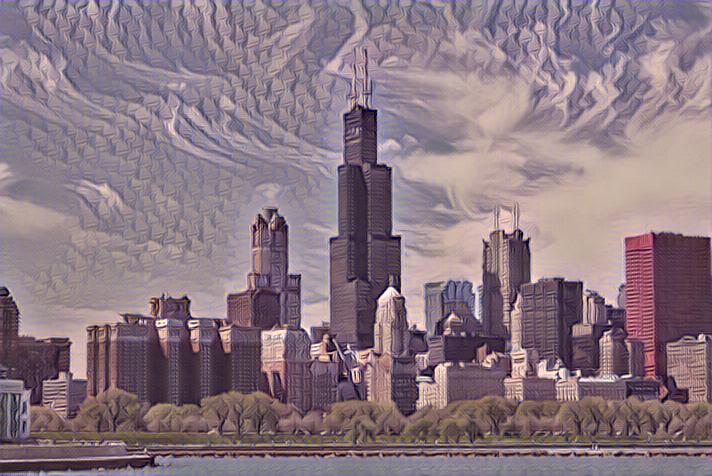}
		\includegraphics[width=0.096\linewidth]{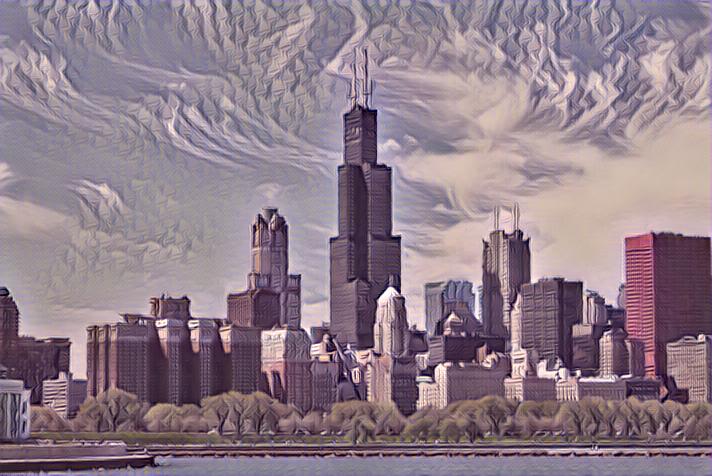}
		\includegraphics[width=0.096\linewidth]{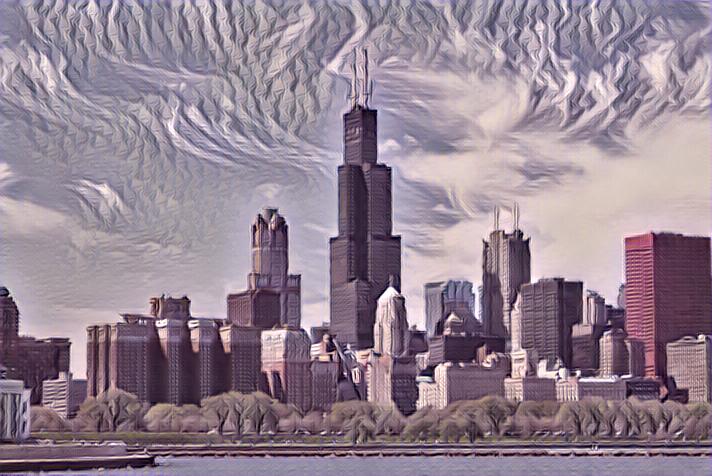}
	}\\
	\end{center}
\vspace{-3mm}
\caption{Comparisons on style morphing between using Dumoulin's algorithm \cite{dumoulin2016learned} (1-st, 3rd and 5-th row) and our method (2-nd, 4-th and 6-th row). Dumoulin's algorithm can indeed create smooth transitions between different styles, but it failed to represent detailed structures. On the contrary, our algorithm can preserve much more detailed structures and create smooth transitions simultaneously.}
\label{style_Com}
\end{figure*}

Fig~\ref{style_Com} compares results between our algorithm and Dumoulin's algorithm \cite{dumoulin2016learned}. As we can see in this experiment, although Dumoulin's algorithm can indeed morph different styles continuously, it failed to represent most of detailed structures. On the contrary, our method can preserve most detailed structures in the images and create smooth transitions between styles.

\subsection{Incremental training}

Observe that in the scenarios that one needs to mix textures/styles with a large number of different relative weights, it can be time-consuming to initialize each optimization process with random noises. We thus proposed to use incremental training to reduce time consumption and create more smooth transitions.
More specifically, our goal is to synthesis $N$ images whose relative weights $\rho$ are equally space in interval $[0,1]$: $\frac{0}{N-1}$, $\frac{1}{N-1}$, .., $\frac{N-2}{N-1}$, $\frac{N-1}{N-1}$. In random training, we simply initialize each image as random noise. In incremental training, we generate these images in sequence from the small weight to the large weight: the first image is initialized with random noise, while the rest of images are initialized with the image synthesized before. The convergence/stop criterion is set to $0.001$ for texture mixing, and the maximum number of iterations is fixed to be $10000$ for each stylish photo.

\begin{figure*}[htb!]
	\begin{center}	
	\TabHeight{
		\includegraphics[width=0.32\linewidth]{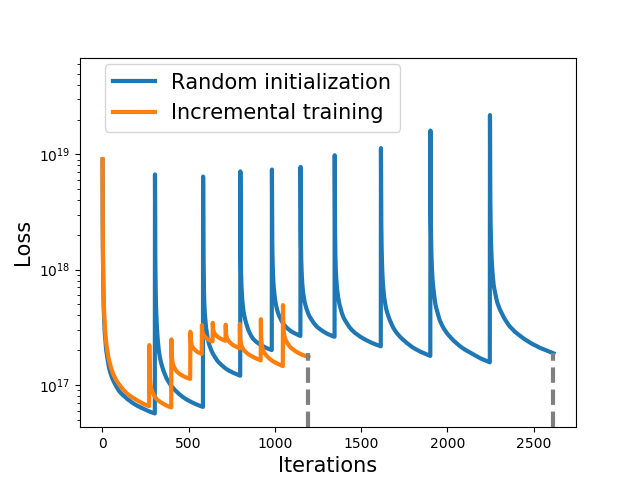}
		\includegraphics[width=0.32\linewidth]{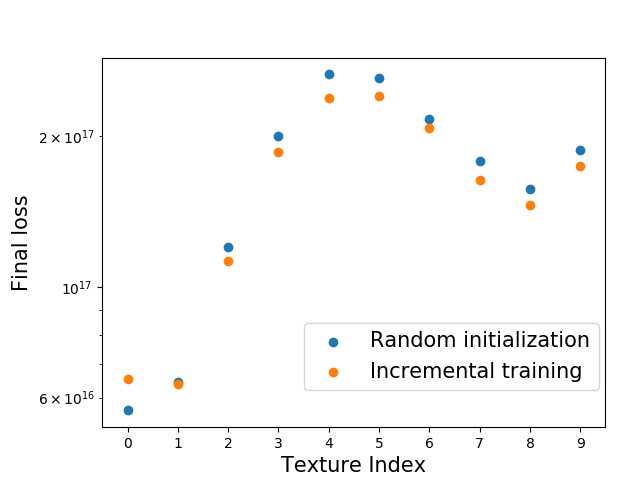}
		\includegraphics[width=0.32\linewidth]{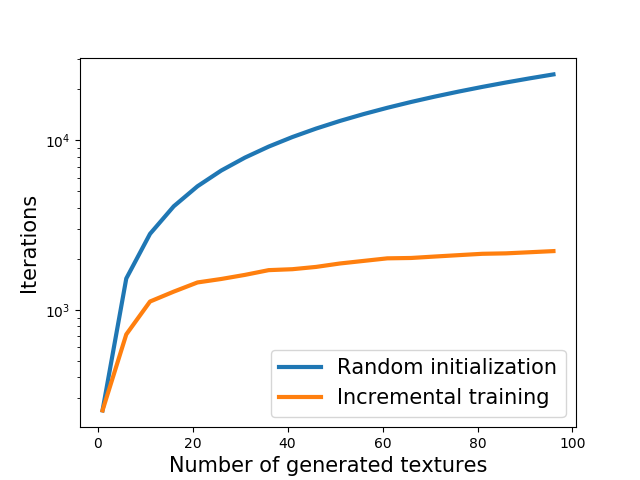}
	}\\
	\TabHeight{
		\includegraphics[width=0.096\linewidth]{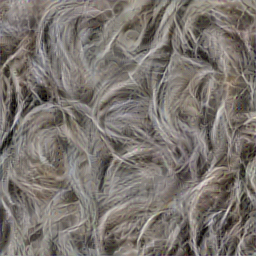}
		\includegraphics[width=0.096\linewidth]{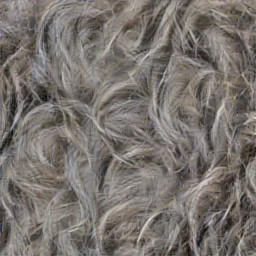}
		\includegraphics[width=0.096\linewidth]{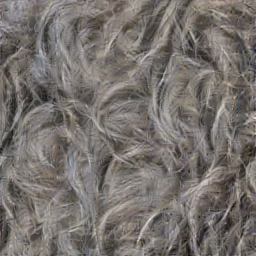}
		\includegraphics[width=0.096\linewidth]{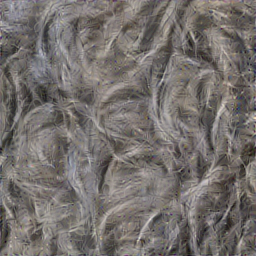}
		\includegraphics[width=0.096\linewidth]{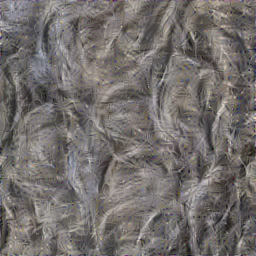}
		\includegraphics[width=0.096\linewidth]{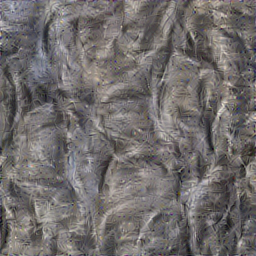}
		\includegraphics[width=0.096\linewidth]{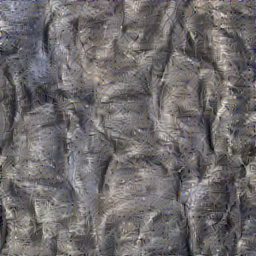}
		\includegraphics[width=0.096\linewidth]{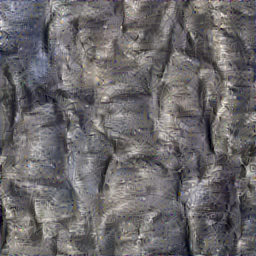}
		\includegraphics[width=0.096\linewidth]{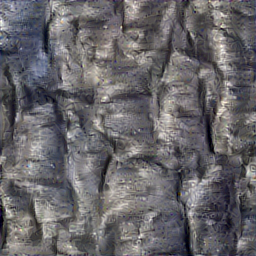}
		\includegraphics[width=0.096\linewidth]{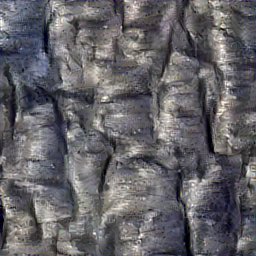}
	}\\
	\TabHeight{
		\includegraphics[width=0.096\linewidth]{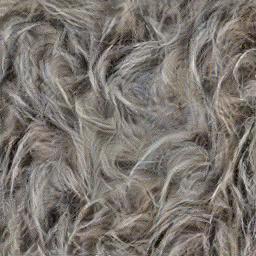}
		\includegraphics[width=0.096\linewidth]{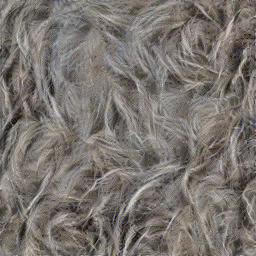}
		\includegraphics[width=0.096\linewidth]{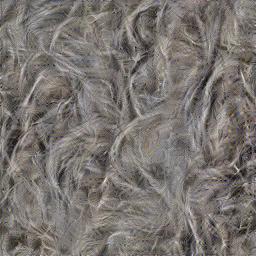}
		\includegraphics[width=0.096\linewidth]{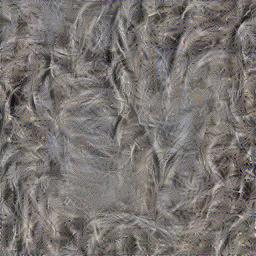}
		\includegraphics[width=0.096\linewidth]{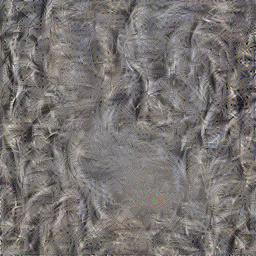}
		\includegraphics[width=0.096\linewidth]{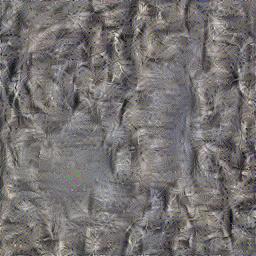}
		\includegraphics[width=0.096\linewidth]{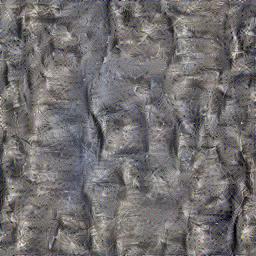}
		\includegraphics[width=0.096\linewidth]{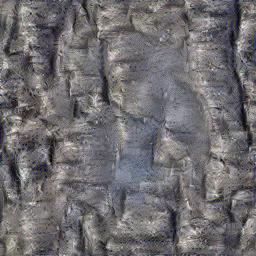}
		\includegraphics[width=0.096\linewidth]{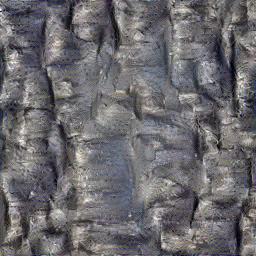}
		\includegraphics[width=0.096\linewidth]{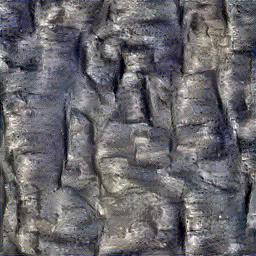}
	}\\
	\end{center}	
\vspace{-3mm}
\caption{Comparison on using incremental and random training for mixing wood and wool textures. Incremental training converges twice faster than random training (Top left), further more, incremental training can also achieve slightly lower loss (Top middle). The difference of converge speed is more obvious when synthesis more images (Top right). Note that, compared with that of using random training (Bottom, 2-nd row), the transitions created by incremental training (Bottom, 1-st row) are more smooth and visually pleasant.}
\label{Incremental}
\end{figure*}

\begin{figure*}[htb!]
	\begin{center}
	\TabHeight{
		\includegraphics[width=0.096\linewidth]{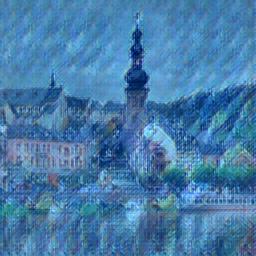}
		\includegraphics[width=0.096\linewidth]{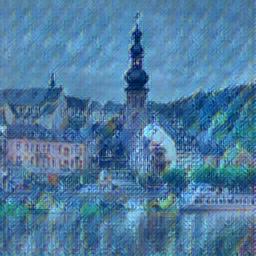}
		\includegraphics[width=0.096\linewidth]{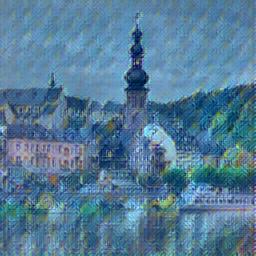}
		\includegraphics[width=0.096\linewidth]{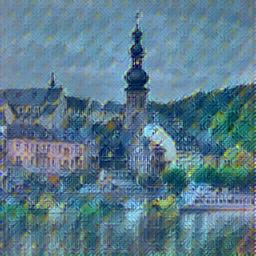}
		\includegraphics[width=0.096\linewidth]{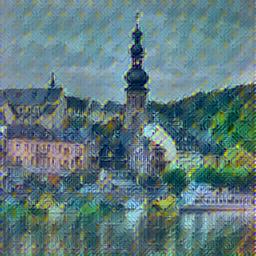}
		\includegraphics[width=0.096\linewidth]{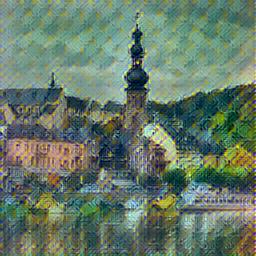}
		\includegraphics[width=0.096\linewidth]{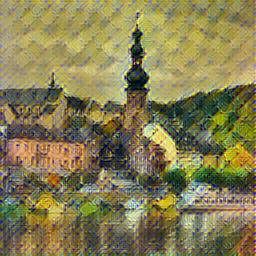}
		\includegraphics[width=0.096\linewidth]{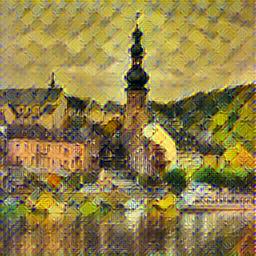}
		\includegraphics[width=0.096\linewidth]{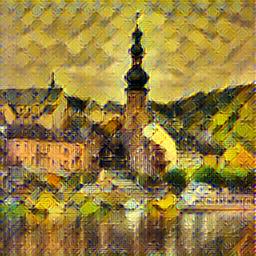}
		\includegraphics[width=0.096\linewidth]{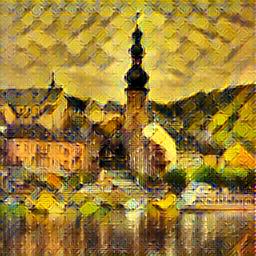}
	}\\
	\TabHeight{
		\includegraphics[width=0.096\linewidth]{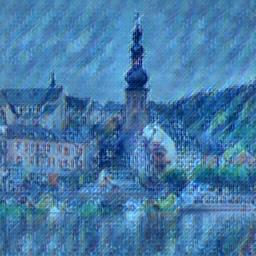}
		\includegraphics[width=0.096\linewidth]{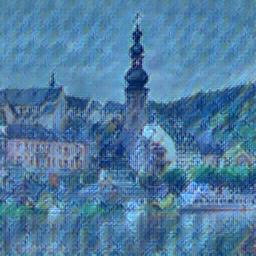}
		\includegraphics[width=0.096\linewidth]{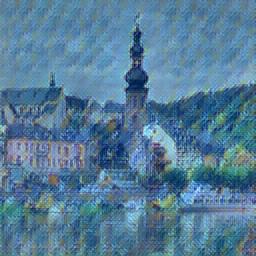}
		\includegraphics[width=0.096\linewidth]{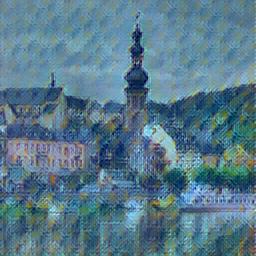}
		\includegraphics[width=0.096\linewidth]{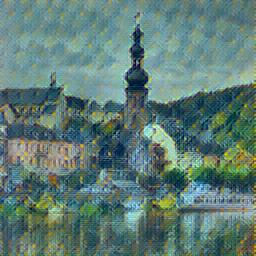}
		\includegraphics[width=0.096\linewidth]{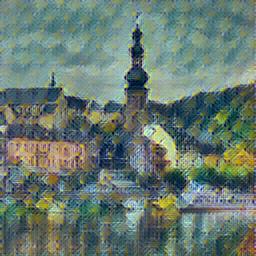}
		\includegraphics[width=0.096\linewidth]{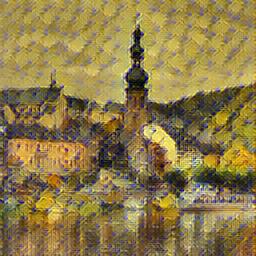}
		\includegraphics[width=0.096\linewidth]{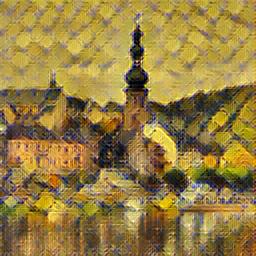}
		\includegraphics[width=0.096\linewidth]{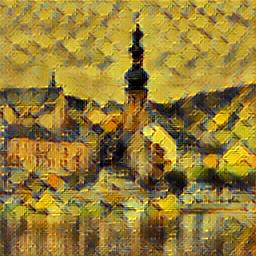}
		\includegraphics[width=0.096\linewidth]{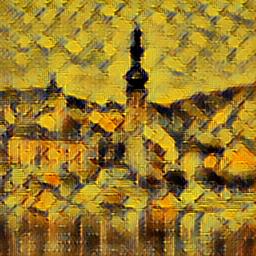}
	}\\
	\end{center}
\vspace{-3mm}
\caption{Comparison on style morphing using incremental training (Bottom, 1-st row) and random training (Bottom, 2-nd row). Notice that the transitions created by incremental training are slightly more smooth.}
\label{Incremental_2}
\end{figure*}

Fig~\ref{Incremental} illustrates the differences between these two training procedures with experiments. For texture mixing, incremental training can speed up the optimization process by offering a better initial point, and also lead to a lower final loss.
As one can see, the difference in time consumption between these two procedures increased with $N$, when $N=100$, incremental training is about one magnitude faster than random training. It is also worth noticing that the adjacent textures created by incremental training is more similar. As a result, the whole transition is more smooth and visually pleasing than random training.
Similar results can be observed in style morphing, see e.g. Fig~\ref{Incremental_2}.

\section{Conclusion}
\label{conclusion}
This paper proposed a novel algorithm to mix textures generated by CNN. We revealed that the statistics used in current CNN based texture models can be written as continuous functions of a Gaussian model, thus, they can be interpolated via Gaussian model. Experimental results have shown that our algorithm excels in mixing high quality textures, and creating mixed styles different from exemplar styles.

There are still some issues need to be further investigated, for example, we notice that the optimization based CNN methods~\cite{gatys2015texture}~\cite{gatys2016image} produce some low level noise. Although in most case one can polish the results with total variation de-noise techniques as in~\cite{johnson2016perceptual}, this problem might be completely overcome by carefully padding the feature maps~\cite{dumoulin2016learned}, or by using upsampling and convolution instead of deconvolution as suggest in~\cite{odena2016deconvolution}, or even retraining CNN weights on the target image.

Another important aspect is the choice of the training set in training feed forward networks. Current researches use the whole ImageNet dataset as the training set, and it is really time consuming to iterate through the whole data set. It is still unclear about the difference in using different training set, e.g. MSCoCo~\cite{lin2014microsoft} and ImageNet, or whether there exist a smaller training set that is effective in style transfer task.
Finally, note that we only described mixing of two given textures/styles, but our algorithm can be extended to mixing more textures/styles without any difficulty.


%
\bibliographystyle{IEEEtran}
\bibliography{egbib}
\end{document}